\documentclass[letterpaper]{article} 
\usepackage[draft]{aaai2026}  
\usepackage{times}  
\usepackage{helvet}  
\usepackage{courier}  
\usepackage[hyphens]{url}  
\usepackage{graphicx} 
\usepackage{natbib}  
\usepackage{caption} 
\usepackage{algorithm}
\usepackage{algorithmic}
\usepackage{amsmath, amsthm, mathtools}
\usepackage{amssymb}   
\usepackage{amsfonts}
\usepackage{booktabs}
\usepackage{multirow}
\usepackage{adjustbox}

\usepackage{newfloat}
\usepackage{listings}
\DeclareCaptionStyle{ruled}{labelfont=normalfont,labelsep=colon,strut=off} 
\floatstyle{ruled}
\newfloat{listing}{tb}{lst}{}
\floatname{listing}{Listing}
\newcommand{\secref}[1]{Section~\ref{#1}}
\newcommand{\figref}[1]{Figure~\ref{#1}}
\newcommand{\tabref}[1]{Table~\ref{#1}}
\newcommand{\eqnref}[1]{Equation~(\ref{#1})}
\newcommand{\appref}[1]{Appendix~(\ref{#1})}
\newcommand{\vk}{\mathbf{k}}
\newcommand{\vv}{\mathbf{v}}
\newcommand{\vq}{\mathbf{q}}
\newcommand{\vu}{\mathbf{u}}
\newcommand{\vp}{\mathbf{p}}

\definecolor{c1}{HTML}{586770}
\definecolor{c4}{HTML}{2a4a67}
\definecolor{c3}{HTML}{6d2a58}
\definecolor{c2}{HTML}{900C3F}

\newcommand{\head}[1]{\vspace{1.7mm}\noindent{{\textcolor{c2}{\bf #1.}}}}

\definecolor{dark2orange}{rgb}{0.9, 0.4, 0.}
\definecolor{dark2purple}{rgb}{0.4, 0.4, 0.8}
\newcommand{\model}{\textsc{Hydra}}
\newcommand{\X}{\mathbf{X}}
\newcommand{\M}{\mathcal{M}}

\newcommand{\x}{\mathbf{x}}
\newcommand{\sequencemodel}{\textsc{EGD-Memory}}
\newcommand{\mb}[1]{\mathbf{#1}}
\newcommand{\undermath}[2]{\underset{#1}{\underbrace{#2}}}

\newtheorem{thm}{Theorem}

\newcommand{\boldres}[1]{{\textbf{{#1}}}}
\newcommand{\secondres}[1]{{\underline{{#1}}}}

\DeclarePairedDelimiter\floor{\lfloor}{\rfloor}

\title{\model{}: Dual Exponentiated Memory for Multivariate Time Series Analysis}
\author{
    Asal Meskin\equalcontrib $^1$, Alireza Mirrokni\equalcontrib $^1$, Ali Najar\equalcontrib $^1$, Ali Behrouz $^2$
}
\affiliations{
    \textsuperscript{\rm 1} Sharif University of Technology \\

    \textsuperscript{\rm 2} Cornell University \\
    \{asal.meskin82, alireza.mirrokni28, ali.najjar82\}@sharif.edu , alibehrouz@cs.cornell.edu
}

\usepackage{bibentry}

\begin{document}

\maketitle

\begin{abstract}
In recent years, effectively modeling multivariate time series has gained significant popularity, mainly due to its wide range of applications, ranging from healthcare to financial markets and energy management. Transformers, MLPs, and linear models as the de facto backbones of modern time series models have shown promising results in single-variant and/or short-term forecasting. These models, however: (1) are permutation equivariant and so lack temporal inductive bias, being less expressive to capture the temporal dynamics; (2) are naturally designed for univariate setup, missing the inter-dependencies of temporal and variate dimensions; and/or (3) are inefficient for Long-term time series modeling. To overcome training and inference efficiency as well as the lack of temporal inductive bias, recently, linear Recurrent Neural Networks (RNNs) have gained attention as an alternative to Transformer-based models. These models, however, are inherently limited to a single sequence, missing inter-variate dependencies, and can propagate errors due to their additive nature. In this paper, we present \model, a by-design two-headed meta in-context memory module that learns how to memorize patterns at test time by prioritizing time series patterns that are more informative about the data. \model{} uses a 2-dimensional recurrence across both time and variate at each step, which is more powerful than mixing methods. Although the 2-dimensional nature of the model makes its training recurrent and non-parallelizable, we present a new 2D-chunk-wise training algorithm that approximates the actual recurrence with $\times 10$ efficiency improvement, while maintaining the effectiveness. Our experimental results on a diverse set of tasks and datasets, including time series forecasting, classification, and anomaly detection show the superior performance of \model{} compared to state-of-the-art baselines. 
\end{abstract}

\section{Introduction} \label{sec:intro}
Modeling multivariate time series has been a crucial task in recent years, mainly due to its connection to understanding and predicting complex systems and its applications in a wide range of
domains such as healthcare~\citep{behrouz2024unsupervised, ivanov1999multifractality}, finance~\citep{gajamannage2023real, pincus2004irregularity}, energy~\citep{zhou2021informer}, transportation~\citep{durango2007time}, and
weather~\citep{allen2025end, price2025probabilistic}. The complex and diverse patterns in multivariate time series data (e.g., seasonal, trend, etc.), its non-stationary nature, and the inter-dependencies among variates make understanding and learning from them considerably challenging. Motivated by the success of Transformers~\citep{vaswani2017attention} in diverse data modalities such as text~\citep{vaswani2017attention, keskar2019ctrl, team2025gemma}, images and videos~\citep{dosovitskiy2020image, liu2021swin}, and graphs~\citep{yun2019graph, muller2024attending} and to address the above challenges, employing Transformer-based architectures for time series has gained popularity~\citep{liu2024itransformer, woo2022etsformer, zhou2021informer, wu2022flowformer, lin2023petformer, chen2024pathformer}.

Transformers, however, have some important inherent characteristics that make their applicability to modeling time series data challenging: (1) Transformers are permutation equivariant and do not have parametric temporal formulation (see \secref{sec:EGD}), lacking natural temporal inductive bias; (2) Their design can cause them to overfit on small-scale datasets like time series data, requiring more sparsity in their weight patterns to be able to adapt to out-of-distribution (OOD) data; and (3) Their quadratic time and memory complexity make them infeasible for many ultra long-term time series modeling. All these limitations have made the applicability of Transformers for time series data challenging. For example, their inherent lack of temporal inductive bias contradicts the causal nature of time series and their complexity that causes overfitting to training data distribution often results in suboptimal performance compared to even simple linear methods~\citep{Zeng2022AreTE}. 

In recent years, there have been significant research efforts to design alternative recurrent architectures that can overcome the efficiency issue of Transformers while maintaining (or improving) their effectiveness~\citep{behrouz2024titans, behrouz2025miras, peng2023rwkv, sun2023retentive, sun2024learning, schlag2021linear}. The recurrent nature of such architectures provide them with a temporal inductive bias and makes them a suitable architectural backbones for time series understanding~\citep{hou2024rwkv, behrouz_chimera_2024, jia2023witran}. Despite this advantage, the additive nature of their recurrence can cause error propagation over time and therefore requires additional careful parametrization or design to achieve a good performance~\citep{jia2023witran, behrouz_chimera_2024}.  Furthermore, a subset of these models (as well as Transformer-based models) overlook the importance of variate dependencies in learning from multivariate time series data~\citep{Zeng2022AreTE, zhang2023effectively, nie2023a}. While several studies aim to design models that take advantage of data mixing across both time and variate~\citep{zhang2022crossformer, li2023mts, wang2022dynamixer}, they might face challenges in real-world tasks \citep{chen2023tsmixer} as these variate dependencies are not always useful in practice; specifically when the target variate is not correlated with other covariates~\citep{chen2023tsmixer}. Therefore, we argue that an effective time series model needs to \emph{adaptively} learn to capture the dependencies of variates over time.

\subsection{Contributions} To overcome the abovementioned challenges, we first present \sequencemodel, a new recurrent sequence model that utilizes a meta in-context memory to learn how to memorize at test time. Following recent surprise-based memory modules~\citep{behrouz2025miras, behrouz2024titans} that prioritize the patterns violating the expectation from the past data, we design \sequencemodel{} so it also memorizes more surprising patterns while promoting the sparsity in the memory weights. This sparsity in the memory is particularly important for time series data as mitigate overfitting to the training set distribution (see our experiments in \secref{sec:experiments} for the details). Building upon \sequencemodel, we present \model, a 2-dimensional variant of \sequencemodel{} that dynamically aggregate the information across both time and variate at each time step. We then show that \model{} is theoretically more powerful than mixing models that separately mix information across time and variates, incorporating the time-mixed and variate-mixed information outside of the hidden space. Despite the expressive power of this model, its design is naturally recurrent and cannot simply be trained in a parallelizable manner. To overcome this issue, we present a novel 2-dimensional chunking algorithm that divides the grid (i.e., multivariate time series) into smaller rectangulars and make the training parallel inside each chunk. In our experimental evaluations, we perform extensive experiments on time series forecasting (ultra-long-term, long-term, and short-term), classification, and anomaly detection tasks. The results indicate the significance of our design and show the superior performance of \model{} compared to state-of-the-art baselines.

\section{Preliminaries and Related Work} \label{sec:preliminaries}
In this section, we briefly provide the notation we use through the paper and then review related studies. More details on the related work and background concepts are provided in \appref{app:rw}.

\subsection{Notation}
We denote a multivariate time series by the matrix $\X = \begin{pmatrix}\x_1 & \dots & \x_V
\end{pmatrix} \in \mathbb{R}^{V \times T \times d_{\text{in}}}$ where $T$ and $V$ are the number of time stamps and variates, and $d_{\text{in}}$ is the feature dimension of the input. We use bold fonts for vectors and bold uppercase letters for matrices. We use $x_{v,t} \in \mathbb{R}^{d_{\text{in}}}$ to refer to the value of the time series in \(v\)-th variate at time \(t\). We follow the literature~\citep{wu2023timesnet, behrouz_chimera_2024} and use ``seasonal patterns'' to refer to repeating and periodic patterns in the data and ``trend patterns'' to refer to general long-term movement in the data. In this paper, we focus on forecasting, classification, and anomaly detection. In forecasting, given the historical time series series \(\X\), the objective is to predict the next \(H\) time steps, producing \(\widetilde{\X} \in \mathbb{R}^{V \times H \times d_{\text{in}}}\). For classification and anomaly detection, the task is to assign a label to the sequence, where anomaly detection is treated as a binary classification problem, distinguishing "normal" from "anomaly" sequences.

\subsection{Autoregressive Process}
Autoregressive models are critical components in time series analysis, predicated on the principle that current observations are influenced by past values, capturing the inherent causal structure within sequential data. For a given integer $p \in \mathbb{N}$ and a vector $\mb{x}_k \in \mathbb{R}^{d}$ representing the observation at time $k$, a standard linear autoregressive model of order $p$, denoted as $\text{AR}(p)$, expresses $\mb{x}_k$ as a linear combination of its $p$ preceding samples:
$$ \mb{x}_k = \sum_{i=1}^{p} \phi_i \mb{x}_{k-i} $$
where $\phi_1, \dots, \phi_p$ are the autoregressive coefficients. This framework can further be extended to accommodate seasonal patterns often observed in time series data. The Seasonal Autoregressive ($\text{SAR}(p, q, s)$) model incorporates dependencies on past observations at seasonal lags. Specifically, it is defined as:
\begin{align}\label{eq:SAR_revised}
    \mb{x}_k = \sum_{i=1}^{p} \phi_i \mb{x}_{k-i} + \sum_{j=1}^{q} \eta_j \mb{x}_{k-js} ,
\end{align}
In this formulation, $s$ denotes the seasonal period (e.g., the number of observations in one seasonal cycle), while $\phi_1, \dots, \phi_p$ (resp. $\eta_1, \dots, \eta_q$) represent the non-seasonal (resp. seasonal) autoregressive coefficients.

\subsection{Traditional Time Series Models}
Modeling complex patterns in time series data has been in the core of research in statistics, mathematics, and computer science. Traditional techniques such as exponential smoothing \citep{winters1960forecasting}, autoregressive integrated moving average (ARIMA) models \citep{bartholomew1971time}, their seasonal counterparts (SARIMA) \citep{bender1994time}, the Box-Jenkins methodology \citep{box1968some} and, in more recent times, state-space models \citep{harvey1990forecasting, aoki2013state} have been the backbone of time series models. Their reliance on hand-crafted features and their limited ability to model non-linear patterns, however, have made their applicability to large-scale complex datasets challenging.

\subsection{Deep Learning-based Time Series Models}
To overcome the limitation of traditional statistical models and with the widespread use of deep learning architectures, Transformers have gained popularity as the backbone of modeling multivariate time series data, primarily due to their effectiveness in modeling complex inter-dependencies among variates and along the temporal dimension \citep{zhou2022fedformer, kitaev2020reformer, zhang2022crossformer, Zeng2022AreTE, zhou2021informer, liu2021pyraformer, wu2021autoformer, ilbert2024unlocking, nie2023a, wu2023timesnet, liu2024itransformer}. Despite the competitive performance of pure Transformer-based architectures, their quadratic time and memory complexity cause critical challenges to use them for Long-term forecasting or on tasks that require modeling Long-term dependencies. To overcome these challenges, recently designing more efficient and effective attention mechanisms that leverage the intrinsic properties of time series data \citep{woo2022etsformer} has gained attention. Concurrently, MLP-based architectures and linear recurrent models~\citep{tolstikhin2021mlp, wang2022dynamixer, behrouz2024titans, sun2024learning, christou2024test} have emerged as promising alternatives to Transformers, demonstrating surprisingly strong performance on various forecasting benchmarks \citep{wu2023timesnet, christou2024test, chen2023tsmixer}. 

Furthermore, convolution-based models that are carefully designed for time series data have shown outstanding performance in various downstream tasks \citep{donghao2024moderntcn, cao2025effectively}. More recently, deep learning models founded on Koopman operator theory have been developed to explicitly address non-linear dynamics within time series \citep{liu2024koopa, wang2023koopman}. However, many existing Koopman-based methods tend to model the dynamics of individual variates separately, which may limit their ability to capture complex cross-variate dependencies. Finally, despite the recent progress in designing naturally 2-dimensional recurrent models~\citep{jia2023witran, behrouz_chimera_2024, behrouz2025leto}, they: (1)  lack data-dependent gating mechanism that filters irrelevant data; (2) are based on recurrent memory that results in a dense representation, potentially overfitting to the training data; (3) rely on attention mechanism that limits the generalization to cases with noisy variates; and/or (3) are based on hebbian-rule memory update that limits the memory management on long forecasting tasks.

\section{\sequencemodel: A Powerful and Robust Sequence Model} \label{sec:EGD}
Next, we present \sequencemodel, a novel recurrent neural architecture that takes advantage of a meta in-context memory module to learn how to memorize at test time (following recent studies on memory design~\citep{sun2024learning, behrouz2024titans, behrouz2025miras}).

\subsection{Neural Memory with Sparse Weights}
In modeling time series data, one of the important factors in designing a powerful model is its ability to capture long-term dependencies. It, however, can be extremely challenging for: (1) Transformers, mainly due to their quadratic time and memory complexity and so their inefficiency when increasing the context length; (2) Recurrent Neural Networks (RNNs), mainly due to their fixed-size memory (i.e., hidden state) that because of compression cannot directly attend to all past time stamps; (3) Convolutions, mainly due to their local nature that fuses the information about each time point with its local neighborhood. We design a long-term memory module that is inherently a recurrent model with a fixed-size memory but: (i) its memory is not a single-matrix and so has more expressive power to compress the information; (ii) its update rule implicitly encourages the compression patterns to be sparse, making the model well-suited for tasks that require adaptability to out-of-distribution data; (iii) has a fast linear-time parallelizable training, making it practically suitable for tasks that require modeling long-term dependencies. 

Following recent advances in sequence modeling~\citep{behrouz2024titans, sun2024learning, behrouz2025miras}, the main idea behind the design of our model is to use a memory module that \emph{learns how to compress the past data into its parameters}. Similar to Transformers~\citep{vaswani2017attention} and modern RNNs~\citep{peng2023rwkv, sun2023retentive, behrouz2024titans}, we build upon the definition of associative memory, in which the memory aims to learn the mapping between a set of keys and a set of values. In practice, these keys and values are different projections of the input sequence, allowing the model to learn underlying patterns between different elements of the sequence. Accordingly, following the literature, we use linear layers to project the input into keys, values, and queries:
\begin{align}\label{eq:keys-values}
    \vk_{t} = W_k \x_{t}, \quad \:\:\: \quad \vv_{t} = W_v \x_{t}, \quad \:\:\: \quad \vq_{t} = W_q \x_{t},
\end{align}
where $W_k, W_v,$ and $W_q$ are learnable matrices. Then, we use the loss function of $\ell(\M_t; \vk_t; \vv_t) = \| \M_t(\vk_t) - \vv_t\|^2_2$ as the objective to measure the predicted mappings between keys and values. Therefore, our memory module $\M_t$ aims to minimize the above objective and so:
\begin{align} \label{eq:objective}
    \M^{*} = \arg \min_{\M} \ell(\M; \vk_t, \vv_t),
\end{align}
for each $t \in \{1, \dots, T\}$. Since $\M$ is a neural network with possible non-nonlinearities, finding the exact optimal solution might not be possible. Therefore, we optimize the objective $\ell(\cdot)$ using an optimization algorithm. A natural choice for the optimization algorithm is gradient descent, a commonly used optimization technique that iteratively updates $\M$ based on the gradient of the objective to converge to an optimal solution. However, the main drawback of such an approach is to overfitting to a local optimal point. To this end, we use a variant of the gradient descent, called Exponentiated Gradient Descent (EGD), that changes the additive nature of the GD to a multiplicative formulation, providing an inherent inductive bias for sparsity in weights~\citep{cornford2024brain}. Given an initial state of memory $\M_0 \in \mathbf{M}$, where $\mathbf{M}$ is an arbitrary class of functions (neural architecture), and input data $\X = \{\x_i\}_{i = 1}^{T}$, EGD optimizes the optimization problem in \eqnref{eq:objective} as follows: 
\begin{align} \label{eq:EGD}
    \M_{t} = \M_{t-1} \odot \exp\left( -\eta_t \nabla \ell(\M_{t-1}; \x_t)\right),
\end{align}
where $\eta_t \in \mathbb{R}^{\geq 0}$ is an adaptive learning rate. Note that this natural inductive bias of EGD for finding sparser solution, which comes from its multiplicative nature~\citep{cornford2024brain}, avoids overfitting and so makes it a powerful optimizer for the scenarios that adaptation to Out-Of-Distribution (OOD) data is a must (e.g., time series data). Later in our experimental results we support the effectiveness of this design, particularly for time series data. 

In this work, we use a simple $\ell_2$ regression loss as the objective, i.e., $\ell(\M; \vk_t, \vv_t) = \| \M(\vk_t) - \vv_t\|^2_2$, where $\vk_t$ and $\vv_t$ are keys and values corresponds to input $\x_t$, obtained from \eqnref{eq:keys-values}. Therefore, using its gradient $\nabla \ell(\M_{t-1}; \vk_t, \vv_t) = (\M_{t-1}(\vk_t) - \vv_t) \vk_t^{\top}$ in \eqnref{eq:EGD}, we have an update rule for the memory $\M$ as:
\begin{align}
    \M_{t} = \M_{t-1} \odot \exp\left( -\eta_t (\M_{t-1}(\vk_t) - \vv_t) \vk_t^{\top} \right).
\end{align}
Assuming that the initial weights are initialized with positive values, we can write the update rule in an additive formulation by taking the $\log$ from both sides of the formulation:
\begin{align}
    \!\!\log\left(\M_{t} \right) = \log\left(\M_{t-1} \right) - \eta_t \M_{t-1}(\vk_t)\vk_t^{\top}\!\! + \eta_t \vv_t \vk_t^{\top}.
\end{align}
Our design of the above recurrent neural network is supported by the fact that memory $\M_t$ is learning the underlying mapping between different components of the input data (e.g., different time points in time series data) by optimizing $\ell(\M; \vk_t, \vv_t) = \| \M(\vk_t) - \vv_t\|^2_2$. Therefore, with enough iterations, we expect the model to converge to an optimal point, in which the memory can map key $\vk_t$ to its corresponding value $\vv_t$, i.e., $\M(\vk_t) \approx \vv_t$. Also, the use of EGD for optimizing the objective guarantees a level of sparsity in the solution, avoiding overfitting and enhancing generalizability (see \secref{sec:experiments} for evaluations). Due to the fixed-size memory state in the above model, we follow \citet{behrouz2024titans} and use a gating mechanism to control the retention of the past memory. That is, we use a learnable input-dependent parameter $\alpha_t \in [0, 1]$ to control the retention from the past state of the memory:
\begin{align} \label{eq:EGD-memory}\tag{\sequencemodel}
    \log\left(\M_{t} \right) = \alpha_t \log\left(\M_{t-1} \right) - \eta_t \M_{t-1}(\vk_t)\vk_t^{\top} + \eta_t \vv_t \vk_t^{\top}.
\end{align}
When $\alpha \rightarrow 1$, the above recurrence retains all the past state of the memory and when $\alpha_t \rightarrow 0$, it erases the past state of the memory, forgetting the past information. 

\subsection{Fast Parallel Training} \label{sec:single-parallel}
Despite the firm theoretical motivations for the design of \eqnref{eq:EGD-memory}, its recurrent formulation is non-linear and cannot be parallelized. The main reason comes from the $\log(\cdot)$ operation that acts as a non-linearity and requires a sequential computation. Following recent advances in hybrid chunk-wise recurrent training algorithms~\citep{sun2024learning, behrouz2025miras}, we divide the sequence $\{\x_{1}, \dots, \x_{T} \}$ with length $T$ into $C$ chunks of size $b = \frac{T}{C}$, each of which is represented by $\mathcal{S}_i = \{ \x_{(i-1)b+1}, \dots, \x_{i b}\}$. Then, for each token $\x_t$ in the $i$-th chunk, we calculate the gradient with respect to the last state of the previous chunk, i.e., $\M_{t'}$, where $t' = \floor{\frac{t}{b}} b$:
\begin{align}\nonumber
    \M_{t} &= \M_{t-1} \odot \exp\left( -\eta_t \nabla \ell(\M_{t'}; \vk_t, \vv_t)\right),\\ \nonumber
    \Rightarrow \: \log\left(\M_{t} \right) &= \alpha_t \log\left(\M_{t-1} \right) - \eta_t \M_{t'}(\vk_t)\vk_t^{\top} + \eta_t \vv_t \vk_t^{\top}.
\end{align}
Note that since we process chunks in a sequential manner, we have computed the state of $\M_{t'}(\cdot)$ before processing the $i$-th chunk, and so all $\vu_t = \M_{t'}(\vk_t)\vk_t^{\top} + \vv_t \vk_t^{\top}$ can be computed at the same time in parallel before starting to process the $i$-th chunk. Therefore, we can revisit the recurrence in each chunk as:
\begin{align}
    \tilde{\M}_t &= \alpha_t \tilde{\M}_{t-1} - \eta_t \vu_t,
\end{align}
where $\tilde{\M}_t = \log\left(\M_{t} \right)$. The above update rule is a linear recurrence and so can be computed in parallel for each chunk using parallel scan or matrix multiplication. For the sake of simplicity, consider the first chunk with initial state of $\tilde{\M}_0$. Then we can simply expand the recurrence as:
\begin{align}
     \tilde{\M}_t &= \alpha_t \tilde{\M}_{t-1} - \eta_t \vu_t \notag \\&= \alpha_1 \dots \alpha_b \tilde{\M}_0 - \sum_{i = 1}^{b} \frac{\alpha_b \dots \alpha_1}{\alpha_i \dots \alpha_1} \times \eta_i \vu_i \notag \\&= \alpha_1 \dots \alpha_b \tilde{\M}_0 - \mathbf{A} \mathbf{E} \mathbf{U},
\end{align}
where $ \mathbf{A} = \texttt{diag}(\frac{\alpha_b \dots \alpha_1}{\alpha_i \dots \alpha_1})$, $\mathbf{E} = \texttt{diag}(\eta_i)$, and $\mathbf{U}$ is the matrix with $i$-th column equal to $\vu_i$, for $i = 1, \dots, b$. Therefore, the above recurrence can be reformulated as matrix multiplication, making it parallelizable for training.

\begin{figure*}
    \centering
    \includegraphics[width=0.75\linewidth]{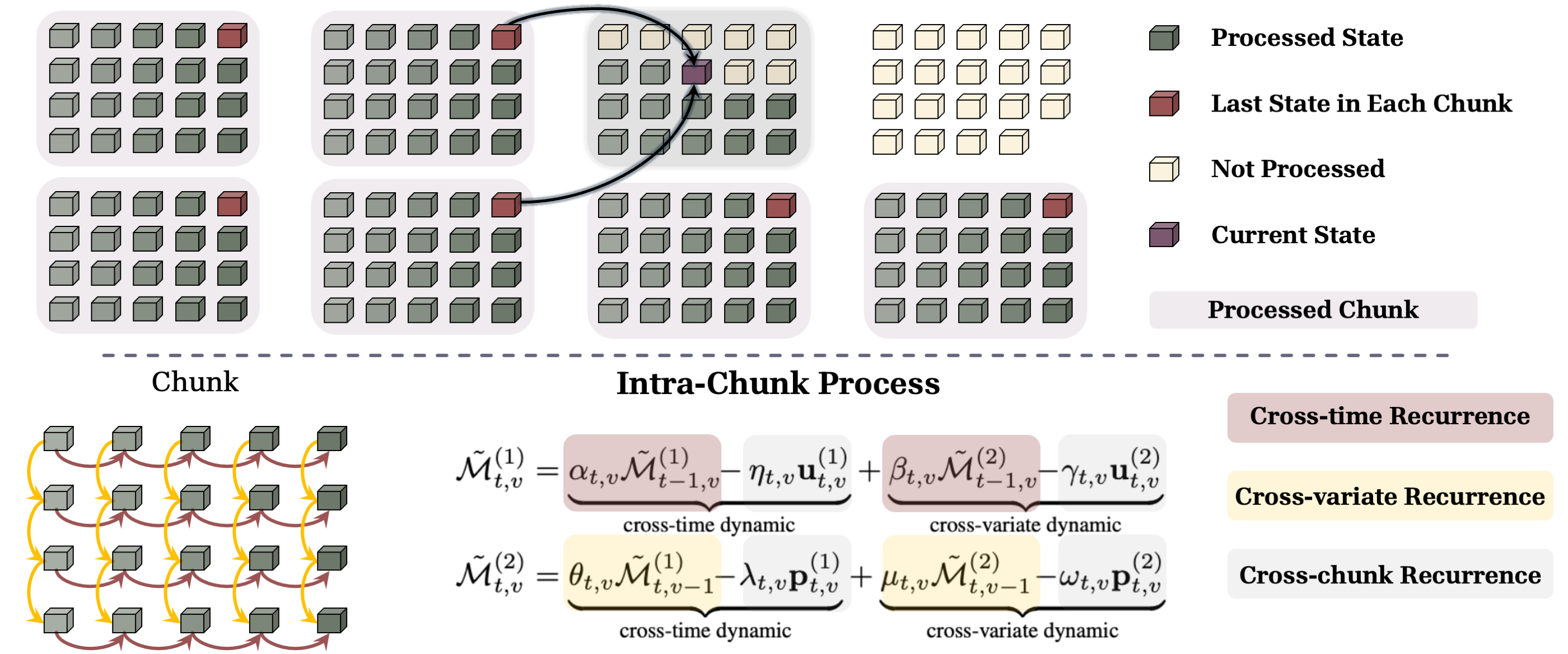}
 \caption{ \model{} at a glance.
\textbf{(a) Chunk-wise processing.} The time series is divided into $b_T \!\times\! b_V$ chunks; processed states are cached (green), the final state is passed forward (maroon), the active state is highlighted (purple), and future states are untouched (cream). 
\textbf{(b) Intra-chunk 2D recurrence.} Within a chunk, computation flows across time (red) and variate (yellow) axes. 
\textbf{(c) Dual memory update.} Closed-form updates for memory heads $\widetilde{\M}^{(1)}$ and $\widetilde{\M}^{(2)}$ capture cross-time (red), cross-variate (yellow), and cross-chunk (grey) dynamics.
}
  \label{fig:model-overview}
\end{figure*}

\section{\model: 2-Dimensional \sequencemodel{} for Multivariate Time Series}\label{sec:hydra}
In the previous section, we presented a novel sequence model motivated by the need for a more expressive memory module and improved memory management (i.e., optimizer) that helps to converge to a sparser solution, which is particularly important for time series data. The main drawback of the described sequence model for multivariate time series data is its univariate nature. That is, it can only learn from a single sequence and misses the rich information of inter-dependencies of variates in multivariate time series data. To this end, in this section, we extend our \sequencemodel{} and present a native 2-dimensional model, called \model{}, that can leverage both cross-time and cross-variate dependencies in the multivariate time series data (see \figref{fig:model-overview}).

\subsection{2-Dimensional Memory Update}
The main idea of \model{} is to use two memory modules $\M^{(1)}$ and $\M^{(2)}$, each of which is responsible for capturing dependencies across time or variates. However, the dynamics of these dependencies can change over time and can affect each other. Therefore, we design a naturally 2-dimensional recurrence that combines the state of each memory over time, capturing the inter-dependencies of dynamics across both time and variate. Let $\X = \begin{pmatrix}
    \x_1 & \dots & \x_V
\end{pmatrix} \in \mathbb{R}^{V \times T \times d_{\text{in}}}$ be the input data. We use $\x_{t,v}$ (resp. $\M_{t,v}$) to refer to the state of the input (resp. memory) at time $t$ in variate $v$. The update rule for each of the memory modules is defined as:
\begin{align} \nonumber
    \log\left(\M^{(1)}_{t, v}\right) &= \undermath{\text{cross-time dynamic}}{ \alpha_{t, v} \log \left(\M^{(1)}_{t-1, v}\right) \!\!- \eta_{t, v} \nabla \ell(\M_{t-1, v}^{(1)}, \x_{t, v})} \! \\ \nonumber &+ \undermath{\text{cross-variate dynamic}}{\beta_{t, v} \log \left(\M^{(2)}_{t-1, v}\right) \!\!- \!\gamma_{t, v} \nabla \ell(\M_{t-1, v}^{(2)}, \x_{t, v})}, \\ \nonumber
    \log\left(\M^{(2)}_{t, v}\right) &= \undermath{\text{cross-time dynamic}}{ \theta_{t, v} \log\left(\M^{(1)}_{t, v-1}\right) \!\!- \! \lambda_{t, v} \nabla \ell(\M_{t, v-1}^{(1)}, \x_{t, v})}  \\ \nonumber &+ \undermath{\text{cross-variate dynamic}}{\mu_{t, v} \log\left(\M^{(2)}_{t, v-1}\right) \!- \! \omega_{t, v} \nabla \ell(\M_{t, v-1}^{(2)}, \x_{t, v})},
\end{align}
where all parameters $\alpha_{t, v}, \eta_{t, v}, \beta_{t, v}, \gamma_{t, v}, \theta_{t, v}, \lambda_{t, v}, \mu_{t, v}$, and $\omega_{t, v}$ are input-dependent learnable parameters. Note that the input dependency here means that these parameters are a function of their corresponding input and can therefore adaptively change based on the context. We use linear layers to obtain such parameters, i.e., $\alpha_{t, v} = W_{\alpha} \x_{v, t}$ and similarly for other parameters of $\beta_{t, v}, \gamma_{t, v}, \theta_{t, v}, \lambda_{t, v}, \mu_{t, v}$, and $\omega_{t, v}$.

\subsubsection{Gating} This formulation provides a flexible information flow between cross-time and cross-variate information. As discussed earlier, while cross-variate dependencies and the flow of information between cross-time and cross-variate dimensions can be very effective in modeling multivariate time series data~\citep{chen2023tsmixer, liu2023itransformer, behrouz_chimera_2024}, this information is not always useful and can degrade performance when the variate is uncorrelated to other variates. Therefore, a powerful model needs to adaptively mix this information. In our design, the parameters  $\alpha_{t, v}, \eta_{t, v}, \beta_{t, v}, \gamma_{t, v}, \theta_{t, v}, \lambda_{t, v}, \mu_{t, v}$, and $\omega_{t, v}$  fully control all the information flow. For example, setting $\beta_{t, v} = \gamma_{t, v} = 0$ can allow the model to filter out cross-variate information when a variate is irrelevant. Or when an input is OOD, if its temporal information is OOD, the model can set $\eta_{t, v} = 0$ and filter its effect on temporal memory, while setting $\gamma_{t, v} \neq 0$ to still use its cross-variate information, if it is useful. Later in our experimental evaluations we show that using this flexible modeling of information flow can significantly improve the performance. Next, we provide two theoretical results to support the design of \model:

\begin{thm}[Expressivity of 2-Dimensional Design] \label{thm:expressiveness}
    \model{} can express full-rank kernels with $\mathcal{O}(1)$ parameters, while linear recurrent neural networks requires at least $\mathcal{O}(N)$ parameters to express a matrix with rank $N$.
\end{thm}

\begin{thm}[Recovering AR process]
    \model{} can recover (seasonal) autoregressive processes in \eqnref{eq:SAR_revised}.
\end{thm}
We provide the details of proofs in \appref{app:proofs}.

\begin{table*}[t]
  \centering
  \renewcommand{\arraystretch}{1.25} 
  \caption{\small Average performance on long term forecasting tasks over four prediction lengths: \{96, 192, 336, 720\}. A lower MAE and MSE indicates a better prediction. The best performance is shown in \boldres{bold}, and the second-best is \secondres{underlined}. Full results are in Appendix.}
  \begin{adjustbox}{max width=\textwidth}
  \begin{tabular}{l cc cc cc cc cc cc cc cc cc cc cc}
    \toprule
    \textbf{Models} & 
    \multicolumn{2}{c}{\textbf{\model{} (Ours)}} & 
    \multicolumn{2}{c}{\textbf{TimeMixer}} & 
    \multicolumn{2}{c}{\textbf{Simba}} & 
    \multicolumn{2}{c}{\textbf{ModernTCN}} &
    \multicolumn{2}{c}{\textbf{iTransformer}} &
    \multicolumn{2}{c}{\textbf{RLinear}} &
    \multicolumn{2}{c}{\textbf{PatchTST}} &
    \multicolumn{2}{c}{\textbf{Crossformer}} &
    \multicolumn{2}{c}{\textbf{TiDE}} &
    \multicolumn{2}{c}{\textbf{TimesNet}} &
    \multicolumn{2}{c}{\textbf{DLinear}} \\ 
    \cmidrule(lr){2-3} \cmidrule(lr){4-5} \cmidrule(lr){6-7} \cmidrule(lr){8-9} 
    \cmidrule(lr){10-11} \cmidrule(lr){12-13} \cmidrule(lr){14-15} \cmidrule(lr){16-17}
    \cmidrule(lr){18-19} \cmidrule(lr){20-21} \cmidrule(lr){21-22} \cmidrule(lr){22-23}
    \textbf{} & \textbf{MSE} & \textbf{MAE} & \textbf{MSE} & \textbf{MAE} & \textbf{MSE} & \textbf{MAE} & \textbf{MSE} & \textbf{MAE} &
    \textbf{MSE} & \textbf{MAE} & \textbf{MSE} & \textbf{MAE} & \textbf{MSE} & \textbf{MAE} &
    \textbf{MSE} & \textbf{MAE} & \textbf{MSE} & \textbf{MAE} & \textbf{MSE} & \textbf{MAE} & \textbf{MSE} & \textbf{MAE} \\ 
    \midrule
    \textbf{ETTm1} & \boldres{0.343} & \boldres{0.372} & 0.381 & 0.385 & 0.383 & 0.396 & \secondres{0.351} & \secondres{0.381} & 0.407 & 0.410 & 0.414 & 0.407 & 0.387 & 0.400 & 0.513 & 0.496 & 0.419 & 0.419 & 0.400 & 0.406 & 0.403 & 0.407 \\ 
    \textbf{ETTm2} & \boldres{0.241} & \boldres{0.298} & 0.275 & 0.323 & 0.271 & 0.327 & \secondres{0.253} & \secondres{0.314} & 0.288 & 0.332 & 0.286 & 0.327 & 0.281 & 0.326 & 0.757 & 0.610 & 0.358 & 0.404 & 0.291 & 0.333 & 0.350 & 0.401 \\
    \textbf{ETTh1} & \boldres{0.394} & \boldres{0.399} & 0.447 & 0.440 & \secondres{0.398} & \secondres{0.409} & {0.404} & 0.420 & 0.454 & 0.447 & 0.446 & 0.434 & 0.469 & 0.454 & 0.529 & 0.522 & 0.541 & 0.507 & 0.458 & 0.450 & 0.456 & 0.452 \\
    \textbf{ETTh2} & \boldres{0.314} & \boldres{0.372} & 0.364 & 0.395 & 0.361 & \secondres{0.377} & \secondres{0.322} & {0.379} & 0.383 & 0.407 & 0.374 & 0.398 & 0.387 & 0.407 & 0.942 & 0.684 & 0.611 & 0.550 & 0.414 & 0.427 & 0.559 & 0.515 \\
    \textbf{Exchange} & \boldres{0.275} & \boldres{0.361} & 0.391 & 0.453 & \secondres{0.298} & \secondres{0.363} & 0.302 & 0.366 & 0.360 & 0.403 & 0.378 & 0.417 & 0.367 & 0.404 & 0.940 & 0.707 & 0.370 & 0.413 & 0.416 & 0.443 & 0.354 & 0.414 \\
    \textbf{Traffic} & 0.401 & \secondres{0.269} & 0.484 & 0.297 & \boldres{0.392} & \boldres{0.264} & \secondres{0.398} & {0.270} & 0.428 & 0.282 & 0.626 & 0.378 & 0.481 & 0.304 & 0.550 & 0.304 & 0.760 & 0.473 & 0.620 & 0.336 & 0.625 & 0.383 \\
    \textbf{Weather} & \boldres{0.210} & \boldres{0.257} & 0.240 & 0.271 & \secondres{0.211} & \secondres{0.258} & {0.224} & {0.264} & 0.258 & 0.278 & 0.272 & 0.291 & 0.259 & 0.281 & 0.259 & 0.315 & 0.271 & 0.320 & 0.259 & 0.287 & 0.265 & 0.317 \\
    \textbf{ECL} & \boldres{0.147} & \boldres{0.244}  & 0.182 & 0.272 & 0.185 & 0.274 & \secondres{0.156} & \secondres{0.253} & 0.178 & 0.270 & 0.219 & 0.298 & 0.205 & 0.290 & 0.244 & 0.334 & 0.251 & 0.344 & 0.192 & 0.295 & 0.212 & 0.300 \\
    \bottomrule
  \end{tabular}
  \end{adjustbox}
  \label{tab:avg_longterm_results}
\end{table*}

\begin{table*}[t]
  \centering
  \renewcommand{\arraystretch}{1.25}
  \caption{\small Average performance on Ultra long-term forecasting tasks (MSE / MAE)}
  \begin{adjustbox}{max width=\textwidth}
  \begin{tabular}{ll cc cc cc cc cc cc cc cc cc}
    \toprule
    \textbf{Dataset} & \textbf{Metric} & 
    \multicolumn{2}{c}{\textbf{\model{}}} &
    \multicolumn{2}{c}{MICN} &
    \multicolumn{2}{c}{TimesNet} &
    \multicolumn{2}{c}{PatchTST} &
    \multicolumn{2}{c}{DLinear} &
    \multicolumn{2}{c}{FiLM} &
    \multicolumn{2}{c}{FEDformer} &
    \multicolumn{2}{c}{Autoformer} &
    \multicolumn{2}{c}{Informer} \\
    
    \cmidrule(lr){3-4} \cmidrule(lr){5-6} \cmidrule(lr){7-8}
    \cmidrule(lr){9-10} \cmidrule(lr){11-12} \cmidrule(lr){13-14}
    \cmidrule(lr){15-16} \cmidrule(lr){17-18} \cmidrule(lr){19-20} 
    
    & & MSE & MAE & MSE & MAE & MSE & MAE & MSE & MAE & MSE & MAE & MSE & MAE & MSE & MAE & MSE & MAE & MSE & MAE \\
    \midrule
    \multirow{3}{*}{ECL}
    & 720–1440  & \boldres{0.4690} & \boldres{0.5267} & 1.0460 & 0.7765 & 0.6119 & 0.5962 & 0.8243 & 0.6704 & 0.4923 & 0.5473 & \secondres{0.4730} & {0.5336} & 0.4833 & 0.5393 & 1.4957 & 0.9533 & 0.5064 & \secondres{0.5317} \\
    & 1440–1440 & \boldres{0.4589} & \boldres{0.5203} & 0.8262 & 1.2207 & 0.5720 & 0.5712 & 0.9053 & 0.7328 & 0.5146 & 0.5615 & \secondres{0.4849} & \secondres{0.5429} & 0.5142 & 0.5571 & 1.7873 & 1.0283 & 0.7247 & 0.6920 \\
    & 1440–2880 & \boldres{0.5811} & \boldres{0.5737} & 2.8936 & 1.3717 & 0.7683 & 0.6846 & 1.1282 & 0.8087 & 0.8355 & 0.7193 & 0.6847 & 0.6493 & 3.9018 & 1.5276 & 1.2867 & 0.8878 & \secondres{0.6152} & \secondres{0.5953} \\
    
    \midrule
    \multirow{3}{*}{Traffic}
    & 720–1440  & \boldres{0.1622} & \secondres{0.2439} & 0.2876 & 0.3916 & 0.1882 & 0.2656 & 0.1904 & 0.2685 & {0.1639} & \boldres{0.2412} & \secondres{0.1638} & 0.2448 & 0.2753 & 0.3650 & 0.3104 & 0.4095 & 0.7614 & 0.6496 \\
    & 1440–1440 & \boldres{0.1499} & 0.2465 & 0.2905 & 0.3923 & 0.2081 & 0.2712 & 0.1917 & 0.2764 & \secondres{0.1590} & \boldres{0.2411} & 0.1602 & \secondres{0.2437} & {0.2848} & 0.3681 & 0.2970 & 0.3999 & 0.7375 & 0.6414 \\
    & 1440–2880 & \boldres{0.1439} & {0.2453} & 0.2823 & 0.3874 & 0.1560 & \boldres{0.2409} & 0.1819 & 0.2761 & \secondres{0.1550} & \secondres{0.2421} & 0.1744 & 0.2693 & 0.2952 & 0.3844 & 0.3035 & 0.3982 & 0.9408 & 0.7618 \\
    
    \midrule
    \multirow{3}{*}{ETTh1}
    & 720–1440  & \boldres{0.1327} & \boldres{0.2856} & 0.4640 & 0.5836 & 0.1391 & \secondres{0.3049} & 0.3708 & 0.4906 & 0.2952 & 0.4370 & 0.2949 & 0.4388 & 0.1768 & 0.3409 & 0.3298 & 0.4741 & 0.1378 & 0.3051 \\
    & 1440–1440 & \boldres{0.1371} & \secondres{0.3108} & 0.5188 & 0.6075 & 0.1404 & \boldres{0.3093} & 0.4475 & 0.5392 & 0.2200 & 0.3714 & 0.3226 & 0.4678 & 0.1928 & 0.3576 & 0.3618 & 0.5507 & \secondres{0.1402} & 0.3192 \\
    & 1440–2880 & \boldres{0.2574} & \secondres{0.3815} & 0.7591 & 0.7215 & 0.2732 & 0.4094 & 0.9617 & 0.8072 & 0.3773 & 0.4794 & 0.3624 & 0.4705 & \secondres{0.2627} & \boldres{0.3754} & 0.3177 & 0.4733 & 0.3495 & 0.4111 \\
    
    \bottomrule
  \end{tabular}
  \end{adjustbox}
  \label{tab:ultralongterm_table_full}
\end{table*}

\subsection{Fast Parallel Training} 
As discussed earlier, even a single memory module with the update rule of \eqnref{eq:EGD-memory} is non-parallelizable, not considering the inter dependencies of two recurrence formula as we discussed for \model. To this end, we extend our chunk-wise training (see \secref{sec:single-parallel}) to 2-dimensional data. In the case of having a sequence as the input, we divide the sequence into some subsequences, however, having  2-dimensional data with 2-dimensional recurrence requires 2-dimensional chunks. To this end, let $\X = \begin{pmatrix}
    \x_1 & \dots & \x_V
\end{pmatrix} \in \mathbb{R}^{V \times T \times d_{\text{in}}}$ be the input data, and $b_T$ and $b_V$ are two integers, indicating the size of chunks across time and variate. We divide $\X$ into $\frac{T}{b_T} \times \frac{V}{b_V}$ non-overlapping rectangulars of size $b_T \times b_V$ and use \[\mathcal{S}_{i, j} = \begin{pmatrix}
    \x_{(i-1)b_T + 1, j b_V} & \dots & \x_{i b_T, j b_V} \\
    \vdots & \ddots & \vdots \\
    \x_{(i-1)b_T + 1, (j-1)b_V + 1 } & \dots & \x_{i b_T, (j-1)b_V + 1 } \\
\end{pmatrix}\] to refer to each chunk. Note that, we index variates in a bottom-up manner. Similar to the univariate case (\secref{sec:single-parallel}), we calculate the gradients with respect to the last state of each rectangular chunk, i.e., $\M^{(1)}_{ib_T, jb_V}$ and $\M^{(2)}_{ib_T, jb_V}$. Hence, we revisit the recurrence in chunk~$\mathcal{S}_{i, j}$~as:
\begin{align} \nonumber
        &\tilde{\M}^{(1)}_{t, v} = \undermath{\text{cross-time dynamic}}{ \alpha_{t, v} \tilde{\M}^{(1)}_{t-1, v} \!\!- \eta_{t, v}  \vu^{(1)}_{t, v}}+ \undermath{\text{cross-variate dynamic}}{\beta_{t, v} \tilde{\M}^{(2)}_{t-1, v} \!\!- \!\gamma_{t, v} \vu^{(2)}_{t, v}}, \\ \nonumber
    &\tilde{\M}^{(2)}_{t, v} = \undermath{\text{cross-time dynamic}}{ \theta_{t, v} \tilde{\M}^{(1)}_{t, v-1} \!\!- \! \lambda_{t, v} \vp^{(1)}_{t, v}} + \undermath{\text{cross-variate dynamic}}{\mu_{t, v} \tilde{\M}^{(2)}_{t, v-1} \!- \! \omega_{t, v} \vp^{(2)}_{t, v}},
\end{align}
where the above auxiliary variables and memory states are defined as \(\vu^{(1)}_{t, v} = \nabla \ell(\M_{t', v'}^{(1)}, \x_{t, v})\), $\vu^{(2)}_{t, v} = \nabla \ell(\M_{t', v'}^{(2)}, \x_{t, v}), \vp^{(1)}_{t'', v''} = \nabla \ell(\M_{t, v}^{(1)}, \x_{t, v}), \vp^{(2)}_{t, v} = \nabla \ell(\M_{t'', v''}^{(2)}, \x_{t, v})$, in which $(t', v') = ((i-1)b_T, jb_v)$ and $(t'', v'') = (ib_T, (j-1)b_V)$. Note that similar to the univariate setup, all parameters above can be computed simultaneously in parallel, before processing this chunk and so the above process, similar to the univariate setup, can be written as a matrix multiplication (see \appref{app:parallel} for the details).

\begin{table*}[th]
  \centering
  \caption{Average performance on short-term forecasting tasks on the M4 dataset. A lower SMAPE, MASE, and OWA indicate better prediction. * is an abbreviation of the "former" term.}
  \begin{small}
  \begin{adjustbox}{max width=\textwidth}
  \renewcommand{\multirowsetup}{\centering}
  \setlength{\tabcolsep}{0.7pt}
  \begin{tabular}{ccc c c c c c c c c c c c c c c}
    \toprule
    \multicolumn{3}{c}{\multirow{1}{*}{\textbf{Models}}} &
    \multicolumn{1}{c}{\textbf{\model{}}} & 
    \multicolumn{1}{c}{\textbf{ModernTCN}} & 
    \multicolumn{1}{c}{\textbf{TimeMixer}} & 
    \multicolumn{1}{c}{\textbf{PatchTST}} & 
    \multicolumn{1}{c}{\textbf{TimesNet}} & 
    \multicolumn{1}{c}{\textbf{N-HiTS}} & 
    \multicolumn{1}{c}{\textbf{N-BEATS$^\ast$}} & 
    \multicolumn{1}{c}{\textbf{ETS$^\ast$}} & 
    \multicolumn{1}{c}{\textbf{LightTS}} & 
    \multicolumn{1}{c}{\textbf{DLinear}} & 
    \multicolumn{1}{c}{\textbf{FED$^\ast$}} & 
    \multicolumn{1}{c}{\textbf{Stationary}} & 
    \multicolumn{1}{c}{\textbf{Auto$^\ast$}} & 
    \multicolumn{1}{c}{\textbf{Pyra$^\ast$}} \\ 
    \midrule
    \multirow{3}{*}{\rotatebox{90}{Weighted}} 
    & \multirow{3}{*}{\rotatebox{90}{Average}} 
    & SMAPE & \boldres{11.651} & \secondres{11.698} & 11.723 & 11.807 & 11.829 & 11.927 & 11.851 & 14.718 & 13.525 & 13.639 & 12.840 & 12.780 & 12.909 & 16.987\\ 
    & & MASE & \boldres{1.535} & \secondres{1.556} & 1.559 & 1.590 & 1.585 & 1.613 & 1.599 & 2.408 & 2.111 & 2.095 & 1.701 & 1.756 & 1.771 & 3.265\\ 
    & & OWA  & \boldres{0.819} & \secondres{0.838} & 0.840 & 0.851 & 0.851 & 0.861 & 0.855 & 1.172 & 1.051 & 1.051 & 0.918 & 0.930 & 0.939 &1.480\\ 
    \bottomrule
  \end{tabular}
  \end{adjustbox}
  \end{small}
  \label{tab:average_shortterm_results}
\end{table*}

\begin{table*}[ht]
    \centering
    \caption{\small Ablation study of \model{} on six datasets (averaged over 5 runs).}
    \scriptsize
    \renewcommand{\arraystretch}{1}
    \begin{adjustbox}{width=0.8\linewidth}
        \begin{tabular}{lcccccc}
                \toprule
                \textbf{Model} & \textbf{ETTh1} & \textbf{ETTh2} & \textbf{ETTm1} & \textbf{ETTm2} & \textbf{Weather} & \textbf{Exchange} \\
                & MSE / MAE & MSE / MAE & MSE / MAE & MSE / MAE & MSE / MAE & MSE / MAE \\
                \midrule
                Full \model{} & \textbf{0.394} / \textbf{0.399} & \textbf{0.314} / \textbf{0.399} & \textbf{0.343} / \textbf{0.372} & \textbf{0.241} / \textbf{0.298} & \textbf{0.210} / \textbf{0.257} & \textbf{0.275} / \textbf{0.364} \\
                w/o cross-variate & 0.461 / 0.443 & 0.457 / 0.481 & 0.414 / 0.429 & 0.318 / 0.369 & 0.277 / 0.294 & 0.335 / 0.410 \\
                w/o input-dependent & 0.479 / 0.464 & 0.392 / 0.421 & 0.407 / 0.410 & 0.341 / 0.370 & 0.258 / 0.278 & 0.360 / 0.403 \\
                w/o Gating & 0.436 / 0.447 & 0.382 / 0.402 & 0.389 / 0.416 & 0.314 / 0.376 & 0.240 / 0.274 & 0.290 / 0.395 \\
                \bottomrule
            \end{tabular}
    \end{adjustbox}
    \label{tab:ablation-study}
\end{table*}

\section{Experiments} \label{sec:experiments}
In this section, we present the experimental setup and results that demonstrate the effectiveness and generality of \model{}. We conduct extensive experiments across three major time series tasks: forecasting (ultra-long-term, long-term, and short-term horizons), classification, and anomaly detection. To evaluate the model’s performance on these tasks, we tested on a variety of benchmark and real-world datasets. All experiments are designed to highlight the model’s ability to generalize and adapt to different temporal patterns while capturing the inter-variate dependencies, showcasing the model’s capacity to effectively learn the relationships between different variates over time.

\subsubsection{Setup}
We evaluate \model{} on 28 publicly available multivariate time series datasets spanning the full range of aforementioned tasks. To contextualize performance, \model{} is compared against over 20 strong and widely adopted baselines, including models based on recurrent and convolutional architectures, multilayer perceptrons, Transformers, and task-specific state-of-the-art methods.
Our experiments follow standard evaluation protocols used in the time series literature. Task-specific evaluation metrics are applied, adhering to conventional interpretations where lower values indicate better performance for error-based metrics and higher values indicate better performance for accuracy-based metrics. Full details on datasets, baselines, and evaluation metrics are provided in \appref{app:experimental_details}.

\subsection{Main Results}

\subsubsection{(Ultra) Long-term Forecasting}
For long-term forecasting, we conduct experiments on a comprehensive suite of benchmark datasets \citep{zhou2021informer} , including Electricity, Traffic, Weather, Exchange Rate, and four settings of ETT datasets: ETTh1, ETTh2, ETTm1, and ETTm2. \model{} achieves state-of-the-art results across all benchmarks for long-term (see \tabref{tab:avg_longterm_results}) and ultra long-term (see \tabref{tab:ultralongterm_table_full}), consistently outperforming Transformer-based and state-space baselines. We attribute these gains to \model{}’s 2D update mechanism,  which simultaneously captures dependencies across time and variates, endowing the model with richer temporal inductive biases than standard Transformer blocks. Moreover, for ultra-long-term forecasting, \model{} employs a learned gating mechanism that dynamically filters out irrelevant or stale historical information, focusing its capacity on the most informative signals and thereby delivering superior forecasting accuracy.

\subsubsection{Short-term Forecasting}
We evaluate \model{} on Short-term forecasting tasks using the M4 benchmark dataset~\citep{godahewa2021monash}. The dataset includes time series from six different domains. As reported in \tabref{tab:average_shortterm_results}, \model{} delivers strong predictive performance across a broad range of tasks, outperforming traditional and state-of-the-art deep learning baselines. Compared to the strongest baseline, \model{} achieves a 2.4\% relative improvement in SMAPE and 1.3\% in MASE. This gain is attributed to inter-dependency  connections and dual-memory architecture.

\subsubsection{Classification}
We evaluate \model{} on ten diverse classification benchmarks from the UEA archive \citep{bagnall2018uea}, covering domains such as biosignals, human activity recognition, and speech. 
As shown in \figref{fig:classificationplot}, \model{} consistently achieves good accuracy across all tasks, outperforming a wide range of baselines, including classical models, convolutional networks, and recent Transformer variants. Notably, \model{} surpasses methods that are tailored for either short or long sequences, demonstrating its robustness and adaptability. This effectiveness stems from \model{}'s chunk-wise recurrence, which enables efficient modeling of local and long-range dependencies.

\subsubsection{Anomaly Detection}
\begin{figure}[ht]
    \centering
    \includegraphics[width=\linewidth]{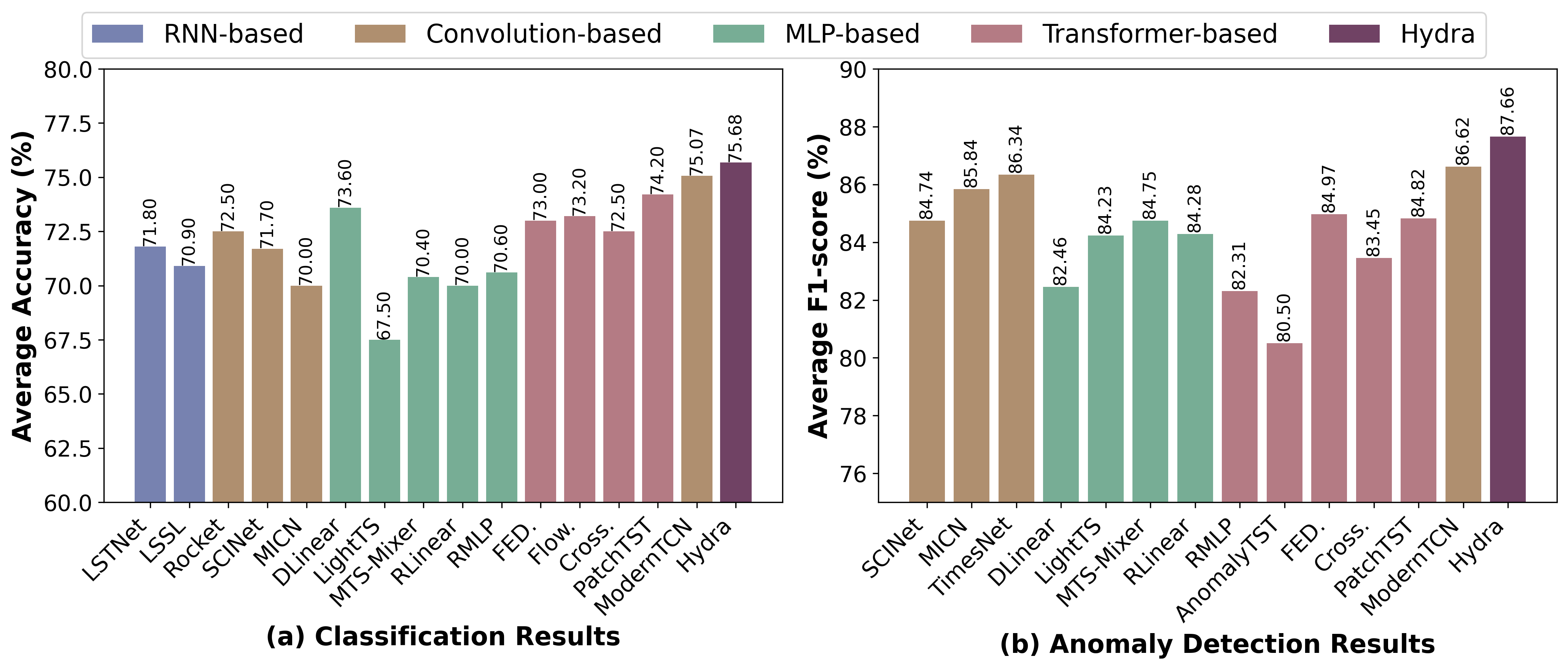}
    \caption{\small Anomaly detection and classification results of \model{} and baselines. Higher accuracy and F1-score indicate better performance. $\cdot$ in the Transformer-based models indicates an abbreviation of the ``former'' term.}
    \label{fig:classificationplot}
\end{figure}

We also evaluate \model{} on five widely-used anomaly detection datasets: SMD \citep{su2019robust}, SWaT \citep{mathur2016swat}, MSL \citep{Hundman_2018}, SMAP \citep{hundman2018detecting}, and PSM \citep{abdulaal2021practical}, which span domains such as service monitoring, space systems, and industrial control.
 \model{} achieves the highest F1 scores across all benchmarks (\figref{fig:classificationplot}), surpassing Transformer-based and classical baselines. These gains come from \model{}'s dual memory design, which tracks information over both time and variates, detecting subtle deviations from expected behavior in both periodic and non-periodic data. By aggregating across chunks, it enhances weak anomaly signals and avoids overfitting to frequent patterns, leading to better generalization on rare or complex anomalies.

\subsection{Ablation Study}
As shown in \tabref{tab:ablation-study}, each component of \model{} plays a critical role, with the full model achieving the lowest errors across all datasets. Removing the cross-variate module limits the model’s ability to learn inter-variable dependencies. Dropping input-dependent parameters reduces adaptability to varying inputs, while removing the gating mechanism weakens effective feature selection. These results clearly demonstrate the importance of each component.

\section{Conclusion}
We present \sequencemodel, a novel native multiplicative recurrent sequence model, that is motivated by the associative memory perspective. Building upon \sequencemodel, we present \model, a native 2-D model that uses two inter-connected memory module each of which responsible to learning the patterns across either time or variates. We present fast parallelizable training process for both models. Our experimental results supports the design of \model, and shows its competitive performance.

\bibliography{main}

\newpage
\appendix
\section{Additional Related Studies}\label{app:rw}

\subsection{Sequence Models}
\subsubsection{Linear and Deep Memory Recurrent Models}
Linear recurrent neural networks (LRNNs) have emerged as efficient alternatives to attention based architectures, offering $O(L)$ time and memory complexity for sequence length $L$ and experiencing renewed interest \citep{tiezzi2024resurgence}. Early work such as RetNet introduced a decaying outer product memory with constant time updates per token, achieving long range retention without quadratic cost \citep{sun2023retentive}. RWKV and state space models further incorporated gating in RNN with outer product recurrence to improve expressiveness \citep{peng2023rwkv, hasani2022liquid, smith2023simplified, behrouz2024mambamixer}. The RWKV6 Eagle model adds dynamic outer product fast weights for even richer memory updates \citep{peng2024eagle}. From a meta learning perspective, DeltaNet interprets linear attention as a fast weight delta rule update \citep{schlag2021delta}, Longhorn casts recurrence as amortized online convex optimization for stable extrapolation \citep{liu2024longhorn}, and more recent test time memorization methods~\citep{behrouz2025miras} such as TTT~\citep{sun2024learning}, Titans, Atlas, and Hope~\citep{behrouz2024titans,behrouz2025atlas, behrouz2025nested} are based on test time parameter updates in a chunk-wise manner for improved efficiency and adaptability. Despite their speed, many RNNs still either process each channel independently or require shallow mixing layers, which can underfit inter variate correlations in multivariate time series. \model{} addresses these gaps with a two headed 2D recurrence that jointly updates across time and variate dimensions, and by equipping each head with a trainable test time memory that selectively retains the most informative patterns for downstream tasks.

\subsubsection{Transformer based Architectures}
Transformers \citep{vaswani2017attention} revolutionized sequence modeling by enabling data driven context weighting through self attention, but their $O(L^2)$ cost limits scalability to long inputs. Sparse and linearized variants, such as Longformer’s local plus global patterns \citep{beltagy2020longformer}, BigBird’s block sparsity \citep{zaheer2020bigbird}, and Performer’s kernelized attention \citep{choromanski2021performer}, reduce this overhead via structured sparsity or random feature maps \citep{katharopoulos2020transformers}. Segment based methods like Transformer XL \citep{dai2019transformerxl} and the Compressive Transformer \citep{rae2020compressive} introduce recurrence over chunk summaries, yet information bottlenecks persist. External memory augmentations — ranging from Neural Turing Machines \citep{graves2014ntm} to retrieval augmented language models like kNN LM \citep{khandelwal2020knn} and RETRO \citep{borgeaud2022retro} — enable post training lookups but often lack seamless on the fly adaptation or efficient indexing for high dimensional multivariate data. In contrast, \model{}’s internal dual head memory mimics the dynamic focus of attention without quadratic complexity or external storage, learning end to end how and when to write or read temporal and variate patterns across arbitrary contexts.
\subsection{Time Series Models}

\subsubsection{Classical Methods}
Classical time series modeling techniques, such as exponential smoothing \citep{winters1960forecasting}, ARIMA \citep{bartholomew1971time}, SARIMA \citep{bender1994time}, the Box-Jenkins method \citep{box1968some}, their multivariate extensions—vector autoregressions (VAR) \citep{sims1980var}, and more recently, state-space models \citep{aoki2013state, harvey1990forecasting}, have long provided interpretable and theoretically grounded approaches. These models serve as strong baselines across forecasting and anomaly detection but rely on assumptions of linearity and stationarity, making them less effective in capturing nonlinear or nonstationary dynamics. Additionally, they often require extensive preprocessing (e.g., differencing, detrending) and manual feature engineering, limiting their scalability and adaptability across complex and evolving datasets.

\subsubsection{Convolutional Approaches}
Convolutional models such as MICN \citep{wang2023micn} and ModernTCN \citep{luo2024moderntcn} utilize dilated or multi-scale kernels to efficiently extract local and mid-range temporal features. TimesNet \citep{wu2023timesnet} transforms time series into two-dimensional period-unfolded tensors and applies Inception-style convolutions to jointly capture seasonal patterns and trends. Other approaches, including ROCKET and MiniROCKET \citep{dempster2020rocket, dempster2021minirocket}, apply random convolutional kernels for classification tasks, achieving strong accuracy with minimal training overhead. Several other convolutional architectures \citep{liu2022scinet, msgnet, tcct, Ismail_Fawaz_2020, NIGAM2025112921} have been proposed to handle diverse modeling challenges. However, convolutional methods often struggle with long-term dependencies and require careful tuning of receptive fields. Their limited capacity to generalize across variable-length patterns and evolving feature dynamics can also hinder performance in more complex anomaly detection or multivariate settings.

\subsubsection{MLP-based Approaches}
MLP-based models such as DLinear \citep{zeng2023dlinear} and TimeMixer \citep{wang2023timemixer} demonstrate that simple linear projections and channel-time mixing can rival Transformer-based methods, especially when combined with trend-seasonality decomposition. FITS, operating entirely in the frequency domain, leverages compact linear layers to interpolate Fourier coefficients \citep{xu2024fits}. Recent work has further improved MLP-based forecasting by incorporating multi-scale and multi-period pattern modeling, knowledge distillation from advanced architectures, and novel MLP designs such as skip and split units \citep{chen2023tsmixer, Unlocking, das2023TiDE, Hong2025, guo2025ramreplaceattentionmlp}. Other MLP variants \citep{Zhang2022LessIM, li2023mts, zhang2023effectively} have been explored across classification settings. Their simplicity enables fast training and inference, but most assume stationary dynamics and lack structural mechanisms for capturing long-term or evolving multivariate relationships. In classification and anomaly detection, this can lead to brittle representations, especially when dealing with nonstationary environments or shifts over time. 

\subsubsection{Transformer-based Time Series Models}
Transformer adaptations for time series, including Informer \citep{zhou2021informer} and Autoformer \citep{wu2021autoformer}, introduce sparse attention mechanisms and decomposition modules to extend temporal context and improve interpretability. PatchTST \citep{nie2023patchtst} segments time series into subseries patches along the temporal dimension to preserve local semantics, while iTransformer \citep{liu2024itransformer} transposes the temporal and variate axes to explicitly model inter-variable dependencies. Transformers have also been applied to classification and anomaly detection. For instance, TST \citep{zerveas2021transformer} refines time series embeddings through self-supervised learning, and Anomaly Transformer \citep{xu2021anomalyTransformer} leverages attention-based discrepancy scores to detect unusual patterns. Some other transformer-based models \citep{deformableTST, wang2024timexer, liu2024unitst, kämäräinen2025minimal} address similar challenges. Despite these strengths, most Transformer models suffer from memory usage that scales quadratically with input length and often require context truncation or downsampling. Additionally, their representations are typically fixed post-training, limiting adaptability to nonstationary or streaming environments. In contrast, \model{} introduces a fixed-size recurrent memory state that enables continuous adaptation, providing robust performance across forecasting, classification, and anomaly detection tasks.

\subsubsection{Multivariate Time Series Modeling}
Handling multivariate time series involves two major challenges: capturing dependencies across variables (inter-variate modeling) and dealing with complex temporal dynamics at multiple resolutions. Recent deep learning models address these issues by employing architectural innovations that combine memory mechanisms, 2D processing, and multi-scale decomposition.

Early approaches relied on vectorizing all variables into a single input vector per time step, which becomes less effective as dimensionality increases. Grid LSTM \citep{kalchbrenner2016gridlongshorttermmemory, kalchbrenner2017neuralmachinetranslationlinear}, BiGridLSTM \citep{fei2018bigradlstm}, and task-specific 2D LSTMs for traffic and industrial data \citep{zhao2017lstmnetwork, liu20202dlstm} pioneered 2D recurrence to explicitly model both temporal and variable relationships. Attention-based models such as dual-attention LSTMs \citep{yu2019two, kim2021dual} and IMV-LSTM \citep{guo2019exploringinterpretablelstmneural} further emphasized variable-specific representations.

Transformer-based models like Crossformer \citep{zhang2023crossformer} and convolutional architectures such as TimesNet \citep{wu2023timesnet} treat time series as 2D matrices, capturing both inter- and intra-period patterns. Additionally, multidimensional recurrent models like Chimera \citep{behrouz_chimera_2024}, WITRAN~\citep{jia2023witran}, and Leto~\citep{behrouz2025leto} extend recurrence to higher dimensions, achieving good results in forecasting, classification, and anomaly detection. These models however, are based on simple Hebbian-rule based recurrence, use linear memory system, are not sparse and so overfit to the training data, and/or requires attention mechanism to perform well.

Complementing these inter-variable strategies, hierarchical and multi-scale architectures further address the challenge of learning at varying temporal resolutions. Models like N-HiTS \citep{challu2022nhits}, Pyraformer \citep{liu2021pyraformer}, Scaleformer \citep{yang2022scaleformer}, and Pathformer \citep{chen2024pathformer} aggregate temporal signals using pyramidal attention or interpolation, capturing both short- and long-term dynamics.

Recent advances expand this direction with models such as the multiscale transformer \citep{naghashi2025multiscale}, which integrates patch-wise temporal modeling with channel-wise representations. Adaptive decomposition frameworks \citep{hu2025adaptivemultiscaledecompositionframework}, multi-resolution architectures \citep{zhang2024multiresolution}, and dual-clustering methods \citep{qiu2024duet} push further in capturing both temporal and variable heterogeneity.

Finally, hybrid CNN-transformer models \citep{hassine2025cnntrans} and MLP-mixer-based decomposition schemes \citep{zhong2024multiscale} demonstrate the strength of combining scale-aware components with flexible inter-variable modeling, yielding robust performance across diverse multivariate tasks.

\section{Proofs of Theorems}\label{app:proofs}
For the prove of both theorems, we show that our \model{} can recover 2D-SSM~\citep{behrouz_chimera_2024, baron2024a}, which are proven to satisfies both of these theorems.  To this end, we need to provide a special instance of our \model{} that is equivalent to these linear 2D recurrent models. For the sake of simplicity, we choose the chunk size as the sequence length, and so all the gradients are with respect to the initial state of the both memories (we let it be $\mathbf{I}$). Therefore, for every $1 \leq t \leq T$, we have:
\begin{align}
    \nabla \ell(\M^{(1)}_{0}; \vk_t, \vv_t) &= (\mathbf{I} \: \vk_t - \vv_t) \vk_t^{\top}  = \vk_t \vk_t^{\top} - \vv_t \vk_t^{\top} = \vu_t \vk_t^{\top},
\end{align}
Now calculating the state of the first memory, we have:
\begin{align}
    \M^{(1)}_{t, v} &= \alpha_{t, v} \M^{(1)}_{t-1, v} - \eta_{t, v} \left( \undermath{\mathbf{u}_t}{(\vk_t  - \vv_t)}\vk_t^{\top} \right) \\&+ \beta_{t, v} \M^{(2)}_{t-1, v} - \gamma_{t, v}  \left(\mathbf{I}\:\: \vk_t \vk_t^{\top} - \vv_t \vk_t^{\top} \right),\notag
\end{align}
where we let $\eta_{t, v} = \gamma_{t, v} = 1$, again for the sake of simplicity. Therefore, it can be written as:
\begin{align}
    &\M^{(1)}_{t, v} = \alpha_{t, v} \M^{(1)}_{t-1, v} - \eta_{t, v} \left( \mathbf{u}_t \vk_t^{\top} \right) + \beta_{t, v} \M^{(2)}_{t-1, v} - \gamma_{t, v}  \mathbf{u}_t,
\end{align}
which is equivalent to the 2-dimensional linear recurrence with diagonal transition matrix. Accordingly, following the prove of \citet{baron2024a}, our \model's recurrence can model full-rank matrix. Also, following the proof of \citet{behrouz_chimera_2024}, this recurrence can recover ARIMA, meaning that a special instance of \model{} can recover autoregressive process.

\section{Parallel Training}\label{app:parallel}
For the sake of clarity, we show the matrix multiplication formulation for only one variate. This formulation can simply be applied with more than one variate. By expanding the recurrence for the first memory, we have:
\begin{align}
    \M^{(1)}_t &= \alpha_t ... \alpha_{t'} \M^{(1)}_{t'} - \sum^{t}_{i=t'} \frac{\alpha_t ... \alpha_{t'}}{\alpha_i ... \alpha_{t'}}\eta_i \undermath{G^{(1)}_t}{\nabla \ell(\M_{t'}; \vk_i, \vv_i)}\notag  \\
    & + \beta ... \beta_{t'} \M^{(2)}_{t'} - \sum^{t}_{i=t'} \frac{\beta_t ... \beta_{t'}}{\beta_i ... \beta_{t'}}\gamma_i \undermath{G^{(2)}_t}{\nabla \ell(\M^{(2)}_{t'}; \vk_i, \vv_i)}\notag \\
    &= \alpha_t ... \alpha_{t'} \M^{(1)}_{t'} + \beta ... \beta_{t'} \M^{(2)}_{t'} - \mathbf{A} \mathbf{E} G^{(1)}_t - \mathbf{B} \mathbf{J} G^{(2)}_t, 
\end{align}
where $\mathbf{A}, \mathbf{E}, \mathbf{B},$ and $\mathbf{J}$ are diagonal matrices with the values of $\frac{\alpha_t ... \alpha_{t'}}{\alpha_i ... \alpha_{t'}}, \frac{\beta_t ... \beta_{t'}}{\beta_i ... \beta_{t'}}, \eta_i,$ and $\gamma_i$.

\section{Experimetal Details}\label{app:experimental_details}

We provide full details of our experimental setup to ensure reproducibility and clarity. First, we describe the datasets we use in our experiments, covering a diverse set of real-world benchmarks. Next, we list all baseline methods against which we compare our model, grouped by their underlying architectures. Finally, we summarize the evaluation metrics employed for each task, highlighting conventions where lower values indicate better performance for error metrics and higher values indicate better outcomes for accuracy-based measures.

\subsection{Datasets}
We evaluate \model{} on 28 publicly available time series datasets, comprising 13 for forecasting (7 for long-term and ultra long-term, and 6 for short-term), 10 for classification, and 5 for anomaly detection \citep{zhou2021informer,godahewa2021monash,bagnall2018uea,su2019robust,mathur2016swat,Hundman_2018,abdulaal2021practical}. These datasets span a wide range of sampling frequencies (from 11\,025 Hz to Yearly), forecasting horizons (6 to 2880 steps), and domains (from electricity and traffic to biosignals and industrial monitoring). \tabref{tab:datasets} provides a summary of each dataset’s dimensionality, train/validation/test splits, sampling frequency, and contextual domain, while \tabref{tab:predlength} lists the exact sequence and prediction lengths for all forecasting benchmarks.

\begin{table*}[t]
\centering
\scriptsize
\setlength{\tabcolsep}{3pt}
\caption{Overview of multivariate time-series datasets used in our experiments.}
\label{tab:datasets}
\begin{adjustbox}{width=\textwidth}
\begin{tabular}{@{}llccc@{}}
\toprule
Tasks & Dataset & Dim & Dataset Size & Information (Frequency) \\
\midrule
\multirow{3}{*}{\shortstack{Forecasting\\(Ultra Long-term)}} 
  & Electricity   & 321 & (18\,317, 2\,633, 5\,261)   & Electricity (hourly) \\
  & Traffic       & 862 & (12\,185, 1\,757, 3\,509)   & Transportation (hourly) \\
  & ETTh1         & 7   & (8\,545, 2\,881, 2\,881)    & Electricity (15 min) \\
\midrule
\multirow{7}{*}{\shortstack{Forecasting\\(Long-term)}} 
  & ETTm1, ETTm2  & 7   & (34\,465, 11\,521, 11\,521) & Electricity (15 min) \\
  & ETTh1, ETTh2  & 7   & (8\,545, 2\,881, 2\,881)    & Electricity (15 min) \\
  & Electricity   & 321 & (18\,317, 2\,633, 5\,261)   & Electricity (hourly) \\
  & Traffic       & 862 & (12\,185, 1\,757, 3\,509)   & Transportation (hourly) \\
  & Weather       & 21  & (36\,792, 5\,271, 10\,540)  & Weather (10 min) \\
  & Exchange      & 8   & (5\,120, 665, 1\,422)       & Exchange rate (daily) \\
\midrule
\multirow{6}{*}{\shortstack{Forecasting\\(Short-term)}} 
  & M4-Yearly     & 1 & (23\,000, –, 23\,000) & Demographic \\
  & M4-Quarterly  & 1 & (24\,000, –, 24\,000) & Finance \\
  & M4-Monthly    & 1 & (48\,000, –, 48\,000) & Industry \\
  & M4-Weekly     & 1 & (359, –, 359)         & Macro \\
  & M4-Daily      & 1 & (4\,227, –, 4\,227)   & Micro \\
  & M4-Hourly     & 1 & (414, –, 414)         & Other \\
\midrule
\multirow{10}{*}{\shortstack{Classification\\(UEA)}} 
  & EthanolConcentration & 3   & (261, –, 263)       & Alcohol industry \\
  & FaceDetection        & 144 & (5\,890, –, 3\,524) & Face (250 Hz) \\
  & Handwriting          & 3   & (150, –, 850)       & Handwriting \\
  & Heartbeat            & 61  & (204, –, 205)       & Heart beat \\
  & JapaneseVowels       & 12  & (270, –, 370)       & Voice \\
  & PEMS-SF              & 963 & (267, –, 173)       & Transportation (daily) \\
  & SelfRegulationSCP1   & 6   & (268, –, 293)       & Health (256 Hz) \\
  & SelfRegulationSCP2   & 7   & (200, –, 180)       & Health (256 Hz) \\
  & SpokenArabicDigits   & 13  & (6\,599, –, 2\,199) & Voice (11 025 Hz) \\
  & UWaveGestureLibrary  & 3   & (120, –, 320)       & Gesture \\
\midrule
\multirow{5}{*}{\shortstack{Anomaly\\Detection}} 
  & SMD  & 38 & (566\,724, 141\,681, 708\,420) & Server machine \\
  & MSL  & 55 & (44\,653, 11\,664, 73\,729)    & Spacecraft \\
  & SMAP & 25 & (108\,146, 27\,037, 427\,617)  & Spacecraft \\
  & SWaT & 51 & (396\,000, 99\,000, 449\,919)  & Infrastructure \\
  & PSM  & 25 & (105\,984, 26\,497, 87\,841)   & Server machine \\
\bottomrule
\end{tabular}
\end{adjustbox}
\end{table*}

\begin{table*}[t]
\centering
\scriptsize
\setlength{\tabcolsep}{3pt}
\caption{Sequence and prediction lengths used for each dataset.}
\label{tab:predlength}
\begin{adjustbox}{width=\textwidth}
\begin{tabular}{@{}llc@{}}
\toprule
Tasks & Dataset & Sequence/Prediction Length \\
\midrule
\multirow{1}{*}{\shortstack{Forecasting (Ultra Long-term)}} 
    & (Same as Table~\ref{tab:datasets})   & \{720/1440, 1440/1440, 1440/2880\} \\
\midrule
\multirow{1}{*}{\shortstack{Forecasting (Long-term)}} 
  & (Same as Table~\ref{tab:datasets}) & \{96/96, 192/192, 336/336, 720/720\} \\
\midrule
\multirow{6}{*}{\shortstack{Forecasting (Short-term)}} 
  & M4-Yearly     & 6/6  \\
  & M4-Quarterly  & 8/8  \\
  & M4-Monthly    & 18/18 \\
  & M4-Weekly     & 13/13 \\
  & M4-Daily      & 14/14 \\
  & M4-Hourly     & 48/48 \\
\midrule
\multirow{10}{*}{\shortstack{Classification\\(UEA)}} 
  & EthanolConcentration & 1\,751/- \\
  & FaceDetection        & 62/-    \\
  & Handwriting          & 152/-   \\
  & Heartbeat            & 405/-   \\
  & JapaneseVowels       & 29/-    \\
  & PEMS-SF              & 144/-   \\
  & SelfRegulationSCP1   & 896/-   \\
  & SelfRegulationSCP2   & 1\,152/- \\
  & SpokenArabicDigits   & 93/-    \\
  & UWaveGestureLibrary  & 315/-   \\
\midrule
\multirow{5}{*}{\shortstack{Anomaly\\Detection}} 
  & SMD  & 100/- \\
  & MSL  & 100/- \\
  & SMAP & 100/- \\
  & SWaT & 100/- \\
  & PSM  & 100/- \\
\bottomrule
\end{tabular}
\end{adjustbox}
\end{table*}

\subsection{Baselines}

We evaluate \model{} against a broad and representative set of baseline models spanning the major paradigms in time series modeling. These include convolutional, MLP-based, state-space, and Transformer-based approaches across forecasting, classification, and anomaly detection tasks.

\head{Long- and Ultra-Long-Term Forecasting}
For long-range forecasting, we consider state-of-the-art models from several architectural families. These include convolutional and MLP-based methods such as DLinear \citep{li2023revisiting}, TimesNet \citep{wu2022timesnet}, SCINet \citep{liu2022scinet}, TimeMixer \citep{wang2023timemixer}, MICN \citep{wang2023micn}, Simba \citep{patro2024simba}, iTransformer \citep{liu2024itransformer}, and TiDE \citep{das2023longterm}. We also evaluate Transformer variants including Informer \citep{zhou2021informer}, Autoformer \citep{wu2021autoformer}, FEDformer \citep{zhou2022fedformer}, FiLM \citep{zhou2022film}, PatchTST \citep{nie2023patchtst}, Crossformer \citep{zhang2022crossformer}, ModernTCN \citep{luo2024moderntcn}, Stationary Transformer \citep{liu2022non}, and DLinear \citep{li2023revisiting} as a simple yet effective linear baseline.

\head{Short-Term Forecasting} 
On the M4 dataset, we follow the standard evaluation suite and include statistical methods such as ETS \citep{winters1960forecasting}, and neural models like N-BEATS \citep{oreshkin2019n} and N-HiTS \citep{challu2022n}. In addition, we test the same set of deep learning models used in long-range forecasting: ModernTCN \citep{luo2024moderntcn}, TimeMixer \citep{wang2023timemixer}, PatchTST \citep{nie2023patchtst}, TimesNet \citep{wu2022timesnet}, LightTS \citep{Zhang2022LessIM}, DLinear \citep{li2023revisiting}, FEDformer \citep{zhou2022fedformer}, Stationary Transformer \citep{liu2022non}, Autoformer \citep{wu2021autoformer}, and Pyraformer \citep{liu2021pyraformer}.

\head{Time Series Classification} 
For classification, we include recurrent models like LSTNet \citep{2018Modeling} and LSSL \citep{sun2024learning}, convolutional methods such as Rocket \citep{Dempster2020ROCKETEF}, SCINet \citep{liu2022scinet}, and MICN \citep{wang2023micn}, as well as MLP-based approaches like DLinear \citep{zeng2023transformers}, LightTS \citep{Zhang2022LessIM}, MTS-Mixer \citep{li2023mts}, RLinear \citep{li2023revisiting}, and RMLP \citep{zhang2023effectively}. Among Transformer-based models, we include FlowFormer \citep{wu2022flowformer}, FEDformer \citep{zhou2022fedformer}, Crossformer \citep{zhang2022crossformer}, PatchTST \citep{nie2023patchtst}, and ModernTCN \citep{luo2024moderntcn}. We further evaluate the following additional baselines in \tabref{tab:full_classification_results}: the classical LSTM \citep{Hochreiter1997LongSM}, vanilla Transformer \citep{vaswani2017attention}, Reformer \citep{kitaev2020reformer}, Informer \citep{zhou2021informer}, Pyraformer \citep{liu2021pyraformer}, Autoformer \citep{wu2021autoformer}, the non-stationary Transformer of \citet{Liu2022NonstationaryTR}, ETSformer \citep{woo2022etsformer},.

\head{Anomaly Detection} 
For anomaly detection, we adopt a similar taxonomy. Convolutional models include SCINet \citep{liu2022scinet}, MICN \citep{wang2023micn}, and TimesNet \citep{wu2022timesnet}; MLP-based models include DLinear \citep{li2023revisiting}, LightTS \citep{Zhang2022LessIM}, MTS-Mixer \citep{li2023mts}, RLinear \citep{li2023revisiting}, and RMLP \citep{zhang2023effectively}; Transformer-based methods include Anomaly Transformer \citep{xu2021anomalyTransformer}, FEDformer \citep{zhou2022fedformer}, Crossformer \citep{zhang2022crossformer}, PatchTST \citep{nie2023patchtst}, and ModernTCN \citep{luo2024moderntcn}.
We further compare against an additional suite of classical and recent baselines (See \tabref{tab:full_anomaly_results}): the original LSTM \citep{Hochreiter1997LongSM}, vanilla Transformer \citep{vaswani2017attention}, LogTrans \citep{2019Enhancing}, TCN \citep{Franceschi2019UnsupervisedSR}, Reformer \citep{kitaev2020reformer}, Informer \citep{zhou2021informer}, Pyraformer \citep{liu2021pyraformer}, Autoformer \citep{wu2021autoformer}, Stationary Transformer \citep{Liu2022NonstationaryTR}, and ETSformer \citep{woo2022etsformer}.

\subsection{Evaluation Metrics}
We evaluate long- and ultra-long-term forecasting using Mean Squared Error (MSE) and Mean Absolute Error (MAE), both measuring average prediction deviation (lower is better). For short-term forecasting, we report Symmetric Mean Absolute Percentage Error (SMAPE), Mean Absolute Scaled Error (MASE), and Overall Weighted Average (OWA), all of which favor lower values. For classification, we use accuracy (higher is better), and for anomaly detection, the F1-score (higher is better).
The metrics are defined as:
\begin{align*} \label{equ:metrics}
    \text{MSE} &= \frac{1}{F}\sum_{i=1}^F (\mathbf{X}_{i} - \widehat{\mathbf{X}}_{i})^2 \\[1.5ex]
    \text{MAE} &= \frac{1}{F} \sum_{i=1}^F \left|\mathbf{X}_{i} - \widehat{\mathbf{X}}_{i}\right| \\[1.5ex]
    \text{SMAPE} &= \frac{200}{F} \sum_{i=1}^F \frac{\left|\mathbf{X}_{i} - \widehat{\mathbf{X}}_{i}\right|}{\left|\mathbf{X}_{i}\right| + \left|\widehat{\mathbf{X}}_{i}\right|} \\[1.5ex]
    \text{MAPE} &= \frac{100}{F} \sum_{i=1}^F \frac{\left|\mathbf{X}_{i} - \widehat{\mathbf{X}}_{i}\right|}{\left|\mathbf{X}_{i}\right|} \\[1.5ex]
    \text{MASE} &= \frac{1}{F} \sum_{i=1}^F \frac{\left|\mathbf{X}_{i} - \widehat{\mathbf{X}}_{i}\right|}{\frac{1}{F-s} \sum_{j=s+1}^{F} \left|\mathbf{X}_{j} - \mathbf{X}_{j-s}\right|} \\[1.5ex]
    \text{OWA} &= \frac{1}{2} \left[ \frac{\text{SMAPE}}{\text{SMAPE}_{\textrm{Naïve2}}} + \frac{\text{MASE}}{\text{MASE}_{\textrm{Naïve2}}} \right] \\[1.5ex]
    \text{Accuracy} &= \frac{\text{TP} + \text{TN}}{\text{TP} + \text{TN} + \text{FP} + \text{FN}} \\[1.5ex]
    \text{F1-score} &= \frac{2 \cdot \text{Precision} \cdot \text{Recall}}{\text{Precision} + \text{Recall}} = \frac{2 \cdot \text{TP}}{2 \cdot \text{TP} + \text{FP} + \text{FN}}
\end{align*}
where $s$ is the periodicity of the data.

\section{Additional Experimental Results}\label{app:exp-results}

\subsection{Long-term Forecasting}
The results are reported in \tabref{tab:fulllongterm}.

\begin{table*}
  \caption{\small Complete experiments on long term forecasting tasks over four prediction lengths: \{96, 192, 336, 720\}. A lower MAE and MSE indicates a better prediction. As a convention for all experimental results, the best performance is shown in \boldres{bold}, and the second-best is \secondres{underlined}. We take the average of 5 separate runs for each prediction length.}\label{tab:fulllongterm}
  \vskip -0.0in
  \vspace{3pt}
  \centering
  \resizebox{1\linewidth}{!}{
  \begin{small}
  \renewcommand{\multirowsetup}{\centering}
  \setlength{\tabcolsep}{1pt}
  \begin{tabular}{c|c |cc | cc | cc |cc| cc|cc|cc|cc|cc|cc|cc|cc|cc|cc|cc}
    \toprule
    \multicolumn{2}{c}{\multirow{2}{*}{}} & 
    \multicolumn{2}{c}{\rotatebox{0}{\scalebox{0.8}{\model{}}}} &
    \multicolumn{2}{c}{\rotatebox{0}{\scalebox{0.8}{{TimeMixer}}}}&
    \multicolumn{2}{c}{\rotatebox{0}{\scalebox{0.8}{{Simba}}}}&
    \multicolumn{2}{c}{\rotatebox{0}{\scalebox{0.8}{{TCN}}}}&
    \multicolumn{2}{c}{\rotatebox{0}{\scalebox{0.8}{{iTransformer}}}} &
    \multicolumn{2}{c}{\rotatebox{0}{\scalebox{0.8}{{RLinear}}}} &
    \multicolumn{2}{c}{\rotatebox{0}{\scalebox{0.8}{PatchTST}}} &
    \multicolumn{2}{c}{\rotatebox{0}{\scalebox{0.8}{Crossformer}}}  &
    \multicolumn{2}{c}{\rotatebox{0}{\scalebox{0.8}{TiDE}}} &
    \multicolumn{2}{c}{\rotatebox{0}{\scalebox{0.8}{{TimesNet}}}} &
    \multicolumn{2}{c}{\rotatebox{0}{\scalebox{0.8}{DLinear}}}&
    \multicolumn{2}{c}{\rotatebox{0}{\scalebox{0.8}{SCINet}}} &
    \multicolumn{2}{c}{\rotatebox{0}{\scalebox{0.8}{FEDformer}}} &
    \multicolumn{2}{c}{\rotatebox{0}{\scalebox{0.8}{Stationary}}} &
    \multicolumn{2}{c}{\rotatebox{0}{\scalebox{0.8}{Autoformer}}} \\
    \multicolumn{2}{c}{} &
    \multicolumn{2}{c}{\scalebox{0.8}{\textbf{(\textcolor{c1}{ours})}}} & 
    \multicolumn{2}{c}{\scalebox{0.8}{{\citeyearpar{wang2023timemixer}}}} & 
    \multicolumn{2}{c}{\scalebox{0.8}{{\citeyearpar{patro2024simba}}}} & 
    \multicolumn{2}{c}{\scalebox{0.8}{\citeyearpar{donghao2024moderntcn}}} & 
    \multicolumn{2}{c}{\scalebox{0.8}{\citeyearpar{liu2024itransformer}}} & 
    \multicolumn{2}{c}{\scalebox{0.8}{\citeyearpar{li2023revisiting}}} & 
    \multicolumn{2}{c}{\scalebox{0.8}{\citeyearpar{nie2023a}}} & 
    \multicolumn{2}{c}{\scalebox{0.8}{\citeyearpar{zhang2022crossformer}}}  & 
    \multicolumn{2}{c}{\scalebox{0.8}{\citeyearpar{das2023longterm}}} & 
    \multicolumn{2}{c}{\scalebox{0.8}{\citeyearpar{wu2023timesnet}}} & 
    \multicolumn{2}{c}{\scalebox{0.8}{\citeyearpar{zeng2023transformers}}}& 
    \multicolumn{2}{c}{\scalebox{0.8}{\citeyearpar{liu2022scinet}}} &
    \multicolumn{2}{c}{\scalebox{0.8}{\citeyearpar{zhou2022fedformer}}} &
    \multicolumn{2}{c}{\scalebox{0.8}{\citeyearpar{liu2022non}}} &
    \multicolumn{2}{c}{\scalebox{0.8}{\citeyearpar{wu2021autoformer}}} \\
    \cmidrule(lr){3-4} \cmidrule(lr){5-6}\cmidrule(lr){7-8} \cmidrule(lr){9-10}\cmidrule(lr){11-12}\cmidrule(lr){13-14} \cmidrule(lr){15-16} \cmidrule(lr){17-18} \cmidrule(lr){19-20} \cmidrule(lr){21-22} \cmidrule(lr){23-24} \cmidrule(lr){25-26} \cmidrule(lr){27-28} \cmidrule(lr){29-30} \cmidrule(lr){30-31} \cmidrule(lr){31-32}
    \multicolumn{2}{c}{}  & \scalebox{0.78}{MSE} & \scalebox{0.78}{MAE} &  \scalebox{0.78}{MSE} & \scalebox{0.78}{MAE} & \scalebox{0.78}{MSE} & \scalebox{0.78}{MAE} &  \scalebox{0.78}{MSE} & \scalebox{0.78}{MAE} & \scalebox{0.78}{MSE} & \scalebox{0.78}{MAE}  & \scalebox{0.78}{MSE} & \scalebox{0.78}{MAE}  & \scalebox{0.78}{MSE} & \scalebox{0.78}{MAE}  & \scalebox{0.78}{MSE} & \scalebox{0.78}{MAE}  & \scalebox{0.78}{MSE} & \scalebox{0.78}{MAE}  & \scalebox{0.78}{MSE} & \scalebox{0.78}{MAE} & \scalebox{0.78}{MSE} & \scalebox{0.78}{MAE} & \scalebox{0.78}{MSE} & \scalebox{0.78}{MAE} & \scalebox{0.78}{MSE} & \scalebox{0.78}{MAE} & \scalebox{0.78}{MSE} & \scalebox{0.78}{MAE} & \scalebox{0.78}{MSE} & \scalebox{0.78}{MAE} \\
    \midrule

    \multirow{5}{*}{{\rotatebox{90}{\scalebox{0.95}{ETTm1}}}}
    &  \scalebox{0.78}{96} &\scalebox{0.78}{0.308} &\scalebox{0.78}{0.338} &  \scalebox{0.78}{0.320} &\scalebox{0.78}{0.357} &\scalebox{0.78}{0.342} & \scalebox{0.78}{0.360} &\scalebox{0.78}{0.292} &\scalebox{0.78}{0.346} &  {\scalebox{0.78}{0.334}} &  {\scalebox{0.78}{0.368}} & \scalebox{0.78}{0.355} & \scalebox{0.78}{0.376} &  {\scalebox{0.78}{0.329}} &  {\scalebox{0.78}{0.367}} & \scalebox{0.78}{0.404} & \scalebox{0.78}{0.426} & \scalebox{0.78}{0.364} & \scalebox{0.78}{0.387} &{\scalebox{0.78}{0.338}} &{\scalebox{0.78}{0.375}} &{\scalebox{0.78}{0.345}} &{\scalebox{0.78}{0.372}} & \scalebox{0.78}{0.418} & \scalebox{0.78}{0.438} &\scalebox{0.78}{0.379} &\scalebox{0.78}{0.419} &\scalebox{0.78}{0.386} &\scalebox{0.78}{0.398} &\scalebox{0.78}{0.505} &\scalebox{0.78}{0.475} \\ 
    & \scalebox{0.78}{192}  &\scalebox{0.78}{0.334} &\scalebox{0.78}{0.361} &  \scalebox{0.78}{0.361} &\scalebox{0.78}{0.381} &\scalebox{0.78}{0.363} & \scalebox{0.78}{0.382} &\scalebox{0.78}{0.332} &\scalebox{0.78}{0.368} & \scalebox{0.78}{0.377} & \scalebox{0.78}{0.391} & \scalebox{0.78}{0.391} & \scalebox{0.78}{0.392} &  {\scalebox{0.78}{0.367}} &  {\scalebox{0.78}{0.385}} & \scalebox{0.78}{0.450} & \scalebox{0.78}{0.451} &\scalebox{0.78}{0.398} & \scalebox{0.78}{0.404} & {\scalebox{0.78}{0.374}} & {\scalebox{0.78}{0.387}}  &{\scalebox{0.78}{0.380}} &{\scalebox{0.78}{0.389}} & \scalebox{0.78}{0.439} & \scalebox{0.78}{0.450}  &\scalebox{0.78}{0.426} &\scalebox{0.78}{0.441} &\scalebox{0.78}{0.459} &\scalebox{0.78}{0.444} &\scalebox{0.78}{0.553} &\scalebox{0.78}{0.496} \\ 
    & \scalebox{0.78}{336} &\scalebox{0.78}{0.343} &\scalebox{0.78}{0.362} &  \scalebox{0.78}{0.390} &\scalebox{0.78}{0.404} &\scalebox{0.78}{0.395} & \scalebox{0.78}{0.405} &\scalebox{0.78}{0.365} &\scalebox{0.78}{0.391} & \scalebox{0.78}{0.426} & \scalebox{0.78}{0.420} & \scalebox{0.78}{0.424} & \scalebox{0.78}{0.415} &  {\scalebox{0.78}{0.399}} &  {\scalebox{0.78}{0.410}} & \scalebox{0.78}{0.532}  &\scalebox{0.78}{0.515} & \scalebox{0.78}{0.428} & \scalebox{0.78}{0.425} & {\scalebox{0.78}{0.410}} & {\scalebox{0.78}{0.411}}  &{\scalebox{0.78}{0.413}} &{\scalebox{0.78}{0.413}} & \scalebox{0.78}{0.490} & \scalebox{0.78}{0.485}  &\scalebox{0.78}{0.445} &\scalebox{0.78}{0.459} &\scalebox{0.78}{0.495} &\scalebox{0.78}{0.464} &\scalebox{0.78}{0.621} &\scalebox{0.78}{0.537} \\ 
    & \scalebox{0.78}{720}  &\scalebox{0.78}{0.387} &\scalebox{0.78}{0.392} &  \scalebox{0.78}{0.454} &\scalebox{0.78}{0.441} &\scalebox{0.78}{0.451} & \scalebox{0.78}{0.437} &\scalebox{0.78}{0.416} &\scalebox{0.78}{0.417} & \scalebox{0.78}{0.491} & \scalebox{0.78}{0.459} & \scalebox{0.78}{0.487} & \scalebox{0.78}{0.450} &  {\scalebox{0.78}{0.454}} &  {\scalebox{0.78}{0.439}} & \scalebox{0.78}{0.666} & \scalebox{0.78}{0.589} & \scalebox{0.78}{0.487} & \scalebox{0.78}{0.461} &{\scalebox{0.78}{0.478}} & {\scalebox{0.78}{0.450}} & {\scalebox{0.78}{0.474}} &{\scalebox{0.78}{0.453}} & \scalebox{0.78}{0.595} & \scalebox{0.78}{0.550}  &\scalebox{0.78}{0.543} &\scalebox{0.78}{0.490} &\scalebox{0.78}{0.585} &\scalebox{0.78}{0.516} &\scalebox{0.78}{0.671} &\scalebox{0.78}{0.561} \\ 
    \cmidrule(lr){2-32}
    & \scalebox{0.78}{Avg}  &\scalebox{0.78}{\boldres{0.343}} &\scalebox{0.78}{\boldres{0.372}} &  \scalebox{0.78}{0.381} &\scalebox{0.78}{0.395} &\scalebox{0.78}{0.383} & \scalebox{0.78}{0.396} &\scalebox{0.78}{\secondres{0.351}} &\scalebox{0.78}{\secondres{0.381}} & \scalebox{0.78}{0.407} & \scalebox{0.78}{0.410} & \scalebox{0.78}{0.414} & \scalebox{0.78}{0.407} &  {\scalebox{0.78}{0.387}} &  {\scalebox{0.78}{0.400}} & \scalebox{0.78}{0.513} & \scalebox{0.78}{0.496} & \scalebox{0.78}{0.419} & \scalebox{0.78}{0.419} & {\scalebox{0.78}{0.400}} & {\scalebox{0.78}{0.406}}  &{\scalebox{0.78}{0.403}} &{\scalebox{0.78}{0.407}} & \scalebox{0.78}{0.485} & \scalebox{0.78}{0.481}  &\scalebox{0.78}{0.448} &\scalebox{0.78}{0.452} &\scalebox{0.78}{0.481} &\scalebox{0.78}{0.456} &\scalebox{0.78}{0.588} &\scalebox{0.78}{0.517} \\ 
    \midrule
    
    \multirow{5}{*}{{\rotatebox{90}{\scalebox{0.95}{ETTm2}}}}
    &  \scalebox{0.78}{96}  &\scalebox{0.78}{0.168} &\scalebox{0.78}{0.245} &  \scalebox{0.78}{0.175} &\scalebox{0.78}{0.258} &\scalebox{0.78}{0.177} & \scalebox{0.78}{0.263} &\scalebox{0.78}{0.166} &\scalebox{0.78}{0.256} &  {\scalebox{0.78}{0.180}} &  {\scalebox{0.78}{0.264}} & \scalebox{0.78}{0.182} & \scalebox{0.78}{0.265} &  {\scalebox{0.78}{0.175}} &  {\scalebox{0.78}{0.259}} & \scalebox{0.78}{0.287} & \scalebox{0.78}{0.366} & \scalebox{0.78}{0.207} & \scalebox{0.78}{0.305} &{\scalebox{0.78}{0.187}} &\scalebox{0.78}{0.267} &\scalebox{0.78}{0.193} &\scalebox{0.78}{{0.292}} & \scalebox{0.78}{0.286} & \scalebox{0.78}{0.377} &\scalebox{0.78}{0.203} &\scalebox{0.78}{0.287} &{\scalebox{0.78}{0.192}} &\scalebox{0.78}{0.274} &\scalebox{0.78}{0.255} &\scalebox{0.78}{0.339} \\ 
    & \scalebox{0.78}{192} &\scalebox{0.78}{0.217} &\scalebox{0.78}{0.269} &  \scalebox{0.78}{0.223} &\scalebox{0.78}{0.299} &\scalebox{0.78}{0.245} & \scalebox{0.78}{0.306} &\scalebox{0.78}{0.222} &\scalebox{0.78}{0.293} & \scalebox{0.78}{0.250} & {\scalebox{0.78}{0.309}} &  {\scalebox{0.78}{0.246}} &  {\scalebox{0.78}{0.304}} &  {\scalebox{0.78}{0.241}} &  {\scalebox{0.78}{0.302}} & \scalebox{0.78}{0.414} & \scalebox{0.78}{0.492} & \scalebox{0.78}{0.290} & \scalebox{0.78}{0.364} &{\scalebox{0.78}{0.249}} &{\scalebox{0.78}{0.309}} &\scalebox{0.78}{0.284} &\scalebox{0.78}{0.362} & \scalebox{0.78}{0.399} & \scalebox{0.78}{0.445} &\scalebox{0.78}{0.269} &\scalebox{0.78}{0.328} &\scalebox{0.78}{0.280} &\scalebox{0.78}{0.339} &\scalebox{0.78}{0.281} &\scalebox{0.78}{0.340} \\ 
    & \scalebox{0.78}{336}  &\scalebox{0.78}{0.270} &\scalebox{0.78}{0.312} &  \scalebox{0.78}{0.299} &\scalebox{0.78}{0.340} &\scalebox{0.78}{0.304} & \scalebox{0.78}{0.343} &\scalebox{0.78}{0.272} &\scalebox{0.78}{0.324} & {\scalebox{0.78}{0.311}} & {\scalebox{0.78}{0.348}} &  {\scalebox{0.78}{0.307}} &  {\scalebox{0.78}{0.342}} &  {\scalebox{0.78}{0.305}} &  {\scalebox{0.78}{0.343}}  & \scalebox{0.78}{0.597} & \scalebox{0.78}{0.542}  & \scalebox{0.78}{0.377} & \scalebox{0.78}{0.422} &{\scalebox{0.78}{0.321}} &{\scalebox{0.78}{0.351}} &\scalebox{0.78}{0.369} &\scalebox{0.78}{0.427} & \scalebox{0.78}{0.637} & \scalebox{0.78}{0.591} &\scalebox{0.78}{0.325} &\scalebox{0.78}{0.366} &\scalebox{0.78}{0.334} &\scalebox{0.78}{0.361} &\scalebox{0.78}{0.339} &\scalebox{0.78}{0.372} \\
    & \scalebox{0.78}{720}  &\scalebox{0.78}{0.351} &\scalebox{0.78}{0.361} &  \scalebox{0.78}{0.391} &\scalebox{0.78}{0.396} &\scalebox{0.78}{0.400} & \scalebox{0.78}{0.399} &\scalebox{0.78}{0.351} &\scalebox{0.78}{0.381} & \scalebox{0.78}{0.412} & \scalebox{0.78}{0.407} &  {\scalebox{0.78}{0.407}} &  {\scalebox{0.78}{0.398}} &  {\scalebox{0.78}{0.402}} &  {\scalebox{0.78}{0.400}} & \scalebox{0.78}{1.730} & \scalebox{0.78}{1.042} & \scalebox{0.78}{0.558} & \scalebox{0.78}{0.524} &{\scalebox{0.78}{0.408}} &{\scalebox{0.78}{0.403}} &\scalebox{0.78}{0.554} &\scalebox{0.78}{0.522} & \scalebox{0.78}{0.960} & \scalebox{0.78}{0.735} &\scalebox{0.78}{0.421} &\scalebox{0.78}{0.415} &\scalebox{0.78}{0.417} &\scalebox{0.78}{0.413} &\scalebox{0.78}{0.433} &\scalebox{0.78}{0.432} \\ 
    \cmidrule(lr){2-32}
    & \scalebox{0.78}{Avg} &\scalebox{0.78}{\boldres{0.241}} &\scalebox{0.78}{\boldres{0.298}} &  \scalebox{0.78}{0.275} &\scalebox{0.78}{0.323} &\scalebox{0.78}{0.271} & \scalebox{0.78}{0.327} &\scalebox{0.78}{0.253} &\scalebox{0.78}{0.314} & {\scalebox{0.78}{0.288}} & {\scalebox{0.78}{0.332}} &  {\scalebox{0.78}{0.286}} &  {\scalebox{0.78}{0.327}} &  {\scalebox{0.78}{0.281}} &  {\scalebox{0.78}{0.326}} & \scalebox{0.78}{0.757} & \scalebox{0.78}{0.610} & \scalebox{0.78}{0.358} & \scalebox{0.78}{0.404} &{\scalebox{0.78}{0.291}} &{\scalebox{0.78}{0.333}} &\scalebox{0.78}{0.350} &\scalebox{0.78}{0.401} & \scalebox{0.78}{0.571} & \scalebox{0.78}{0.537} &\scalebox{0.78}{0.305} &\scalebox{0.78}{0.349} &\scalebox{0.78}{0.306} &\scalebox{0.78}{0.347} &\scalebox{0.78}{0.327} &\scalebox{0.78}{0.371} \\ 
    \midrule
    
    \multirow{5}{*}{\rotatebox{90}{{\scalebox{0.95}{ETTh1}}}}
    &  \scalebox{0.78}{96}  &\scalebox{0.78}{0.358} &\scalebox{0.78}{0.379} &  \scalebox{0.78}{0.375} &\scalebox{0.78}{0.400} &\scalebox{0.78}{0.379} & \scalebox{0.78}{0.395} &\scalebox{0.78}{0.368} &\scalebox{0.78}{0.394} & {\scalebox{0.78}{0.386}} & {\scalebox{0.78}{0.405}} & \scalebox{0.78}{0.386} &  {\scalebox{0.78}{0.395}} & \scalebox{0.78}{0.414} & \scalebox{0.78}{0.419} & \scalebox{0.78}{0.423} & \scalebox{0.78}{0.448} & \scalebox{0.78}{0.479}& \scalebox{0.78}{0.464}  & {\scalebox{0.78}{0.384}} &{\scalebox{0.78}{0.402}} & \scalebox{0.78}{0.386} & {\scalebox{0.78}{0.400}} & \scalebox{0.78}{0.654} & \scalebox{0.78}{0.599} & {\scalebox{0.78}{0.376}} &\scalebox{0.78}{0.419} &\scalebox{0.78}{0.513} &\scalebox{0.78}{0.491} &\scalebox{0.78}{0.449} &\scalebox{0.78}{0.459}  \\ 
    & \scalebox{0.78}{192}  &\scalebox{0.78}{0.383} &\scalebox{0.78}{0.392} &  \scalebox{0.78}{0.429} &\scalebox{0.78}{0.421} &\scalebox{0.78}{0.432} & \scalebox{0.78}{0.424} &\scalebox{0.78}{0.405} &\scalebox{0.78}{0.413} & \scalebox{0.78}{0.441} & \scalebox{0.78}{0.436} & {\scalebox{0.78}{0.437}} &  {\scalebox{0.78}{0.424}} & \scalebox{0.78}{0.460} & \scalebox{0.78}{0.445} & \scalebox{0.78}{0.471} & \scalebox{0.78}{0.474}  & \scalebox{0.78}{0.525} & \scalebox{0.78}{0.492} & {\scalebox{0.78}{0.436}} & {\scalebox{0.78}{0.429}}  &{\scalebox{0.78}{0.437}} &{\scalebox{0.78}{0.432}} & \scalebox{0.78}{0.719} & \scalebox{0.78}{0.631} & {\scalebox{0.78}{0.420}} &\scalebox{0.78}{0.448} &\scalebox{0.78}{0.534} &\scalebox{0.78}{0.504} &\scalebox{0.78}{0.500} &\scalebox{0.78}{0.482} \\ 
    & \scalebox{0.78}{336}  &\scalebox{0.78}{0.457} &\scalebox{0.78}{0.451} &  \scalebox{0.78}{0.484} &\scalebox{0.78}{0.458} &\scalebox{0.78}{0.473} & \scalebox{0.78}{0.443} &\scalebox{0.78}{0.391} &\scalebox{0.78}{0.412} & {\scalebox{0.78}{0.487}} &  {\scalebox{0.78}{0.458}} &  {\scalebox{0.78}{0.479}} &  {\scalebox{0.78}{0.446}} & \scalebox{0.78}{0.501} & \scalebox{0.78}{0.466} & \scalebox{0.78}{0.570} & \scalebox{0.78}{0.546} & \scalebox{0.78}{0.565} & \scalebox{0.78}{0.515} &\scalebox{0.78}{0.491} &\scalebox{0.78}{0.469} &{\scalebox{0.78}{0.481}} & {\scalebox{0.78}{0.459}} & \scalebox{0.78}{0.778} & \scalebox{0.78}{0.659} & {\scalebox{0.78}{0.459}} &{\scalebox{0.78}{0.465}} &\scalebox{0.78}{0.588} &\scalebox{0.78}{0.535} &\scalebox{0.78}{0.521} &\scalebox{0.78}{0.496} \\ 
    & \scalebox{0.78}{720}  &\scalebox{0.78}{0.429} &\scalebox{0.78}{0.424} &  \scalebox{0.78}{0.498} &\scalebox{0.78}{0.482} &\scalebox{0.78}{0.483} & \scalebox{0.78}{0.469} &\scalebox{0.78}{0.450} &\scalebox{0.78}{0.461} & {\scalebox{0.78}{0.503}} & {\scalebox{0.78}{0.491}} &  {\scalebox{0.78}{0.481}} & 
     {\scalebox{0.78}{0.470}} &  {\scalebox{0.78}{0.500}} &  {\scalebox{0.78}{0.488}} & \scalebox{0.78}{0.653} & \scalebox{0.78}{0.621} & \scalebox{0.78}{0.594} & \scalebox{0.78}{0.558} &\scalebox{0.78}{0.521} &{\scalebox{0.78}{0.500}} &\scalebox{0.78}{0.519} &\scalebox{0.78}{0.516} & \scalebox{0.78}{0.836} & \scalebox{0.78}{0.699} &{\scalebox{0.78}{0.506}} &{\scalebox{0.78}{0.507}} &\scalebox{0.78}{0.643} &\scalebox{0.78}{0.616} &{\scalebox{0.78}{0.514}} &\scalebox{0.78}{0.512}  \\ 
    \cmidrule(lr){2-32}
    & \scalebox{0.78}{Avg} &\scalebox{0.78}{\boldres{0.394}} &\scalebox{0.78}{\boldres{0.399}} &  \scalebox{0.78}{0.447} &\scalebox{0.78}{0.440} &\scalebox{0.78}{0.441} & \scalebox{0.78}{0.432} &\scalebox{0.78}{\secondres{0.404}} &\scalebox{0.78}{\secondres{0.420}} & {\scalebox{0.78}{0.454}} &  {\scalebox{0.78}{0.447}} &  {\scalebox{0.78}{0.446}} &  {\scalebox{0.78}{0.434}} & \scalebox{0.78}{0.469} & \scalebox{0.78}{0.454} & \scalebox{0.78}{0.529} & \scalebox{0.78}{0.522} & \scalebox{0.78}{0.541} & \scalebox{0.78}{0.507} &\scalebox{0.78}{0.458} &{\scalebox{0.78}{0.450}} &{\scalebox{0.78}{0.456}} &{\scalebox{0.78}{0.452}} & \scalebox{0.78}{0.747} & \scalebox{0.78}{0.647} & {\scalebox{0.78}{0.440}} &\scalebox{0.78}{0.460} &\scalebox{0.78}{0.570} &\scalebox{0.78}{0.537} &\scalebox{0.78}{0.496} &\scalebox{0.78}{0.487}  \\ 
    \midrule

    \multirow{5}{*}{\rotatebox{90}{\scalebox{0.95}{ETTh2}}}
    &  \scalebox{0.78}{96}  &\scalebox{0.78}{0.248} &\scalebox{0.78}{0.321} &  \scalebox{0.78}{0.289} &\scalebox{0.78}{0.341} &\scalebox{0.78}{0.290} & \scalebox{0.78}{0.339} &\scalebox{0.78}{0.263} &\scalebox{0.78}{0.332} &  {\scalebox{0.78}{0.297}} & {\scalebox{0.78}{0.349}} &  {\scalebox{0.78}{0.288}} &  {\scalebox{0.78}{0.338}} & {\scalebox{0.78}{0.302}} &  {\scalebox{0.78}{0.348}} & \scalebox{0.78}{0.745} & \scalebox{0.78}{0.584} &\scalebox{0.78}{0.400} & \scalebox{0.78}{0.440}  & {\scalebox{0.78}{0.340}} & {\scalebox{0.78}{0.374}} &{\scalebox{0.78}{0.333}} &{\scalebox{0.78}{0.387}} & \scalebox{0.78}{0.707} & \scalebox{0.78}{0.621}  &\scalebox{0.78}{0.358} &\scalebox{0.78}{0.397} &\scalebox{0.78}{0.476} &\scalebox{0.78}{0.458} &\scalebox{0.78}{0.346} &\scalebox{0.78}{0.388} \\ 
    & \scalebox{0.78}{192}  &\scalebox{0.78}{0.305} &\scalebox{0.78}{0.368} &  \scalebox{0.78}{0.372} &\scalebox{0.78}{0.392} &\scalebox{0.78}{0.373} & \scalebox{0.78}{0.390} &\scalebox{0.78}{0.320} &\scalebox{0.78}{0.374} &  {\scalebox{0.78}{0.380}} &  {\scalebox{0.78}{0.400}} &  {\scalebox{0.78}{0.374}} &  {\scalebox{0.78}{0.390}} &{\scalebox{0.78}{0.388}} & {\scalebox{0.78}{0.400}} & \scalebox{0.78}{0.877} & \scalebox{0.78}{0.656} & \scalebox{0.78}{0.528} & \scalebox{0.78}{0.509} & {\scalebox{0.78}{0.402}} & {\scalebox{0.78}{0.414}} &\scalebox{0.78}{0.477} &\scalebox{0.78}{0.476} & \scalebox{0.78}{0.860} & \scalebox{0.78}{0.689} &{\scalebox{0.78}{0.429}} &{\scalebox{0.78}{0.439}} &\scalebox{0.78}{0.512} &\scalebox{0.78}{0.493} &\scalebox{0.78}{0.456} &\scalebox{0.78}{0.452} \\ 
    & \scalebox{0.78}{336}  &\scalebox{0.78}{0.305} &\scalebox{0.78}{0.378} &  \scalebox{0.78}{0.386} &\scalebox{0.78}{0.414} &\scalebox{0.78}{0.376} & \scalebox{0.78}{0.406} &\scalebox{0.78}{0.313} &\scalebox{0.78}{0.376} &  {\scalebox{0.78}{0.428}} &  {\scalebox{0.78}{0.432}} &  {\scalebox{0.78}{0.415}} &  {\scalebox{0.78}{0.426}} &  {\scalebox{0.78}{0.426}} & {\scalebox{0.78}{0.433}}& \scalebox{0.78}{1.043} & \scalebox{0.78}{0.731} & \scalebox{0.78}{0.643} & \scalebox{0.78}{0.571}  & {\scalebox{0.78}{0.452}} & {\scalebox{0.78}{0.452}} &\scalebox{0.78}{0.594} &\scalebox{0.78}{0.541} & \scalebox{0.78}{1.000} &\scalebox{0.78}{0.744} &\scalebox{0.78}{0.496} &\scalebox{0.78}{0.487} &\scalebox{0.78}{0.552} &\scalebox{0.78}{0.551} &{\scalebox{0.78}{0.482}} &\scalebox{0.78}{0.486}\\ 
    & \scalebox{0.78}{720}  &\scalebox{0.78}{0.374} &\scalebox{0.78}{0.429} &  \scalebox{0.78}{0.412} &\scalebox{0.78}{0.434} &\scalebox{0.78}{0.407} & \scalebox{0.78}{0.431} &\scalebox{0.78}{0.392} &\scalebox{0.78}{0.433} &  {\scalebox{0.78}{0.427}} &  {\scalebox{0.78}{0.445}} &  {\scalebox{0.78}{0.420}} &  {\scalebox{0.78}{0.440}} & {\scalebox{0.78}{0.431}} & {\scalebox{0.78}{0.446}} & \scalebox{0.78}{1.104} & \scalebox{0.78}{0.763} & \scalebox{0.78}{0.874} & \scalebox{0.78}{0.679} & {\scalebox{0.78}{0.462}} & {\scalebox{0.78}{0.468}} &\scalebox{0.78}{0.831} &\scalebox{0.78}{0.657} & \scalebox{0.78}{1.249} & \scalebox{0.78}{0.838} &{\scalebox{0.78}{0.463}} &{\scalebox{0.78}{0.474}} &\scalebox{0.78}{0.562} &\scalebox{0.78}{0.560} &\scalebox{0.78}{0.515} &\scalebox{0.78}{0.511} \\ 
    \cmidrule(lr){2-32}
    & \scalebox{0.78}{Avg}  &\scalebox{0.78}{\boldres{0.309}} &\scalebox{0.78}{0.365} &  \scalebox{0.78}{0.364} &\scalebox{0.78}{0.395} &\scalebox{0.78}{0.361} & \scalebox{0.78}{\boldres{0.377}} &\scalebox{0.78}{\secondres{0.322}} &\scalebox{0.78}{\secondres{0.379}} &  {\scalebox{0.78}{0.383}} &  {\scalebox{0.78}{0.407}} &  {\scalebox{0.78}{0.374}} &  {\scalebox{0.78}{0.398}} & {\scalebox{0.78}{0.387}} & {\scalebox{0.78}{0.407}} & \scalebox{0.78}{0.942} & \scalebox{0.78}{0.684} & \scalebox{0.78}{0.611} & \scalebox{0.78}{0.550}  &{\scalebox{0.78}{0.414}} &{\scalebox{0.78}{0.427}} &\scalebox{0.78}{0.559} &\scalebox{0.78}{0.515} & \scalebox{0.78}{0.954} & \scalebox{0.78}{0.723} &\scalebox{0.78}{{0.437}} &\scalebox{0.78}{{0.449}} &\scalebox{0.78}{0.526} &\scalebox{0.78}{0.516} &\scalebox{0.78}{0.450} &\scalebox{0.78}{0.459} \\ 
    \midrule

    \multirow{5}{*}{\rotatebox{90}{{\scalebox{0.95}{Exchange}}}}
    &  \scalebox{0.78}{96}  &\scalebox{0.78}{0.075} &\scalebox{0.78}{0.203} &  \scalebox{0.78}{0.090} &\scalebox{0.78}{0.235} &\scalebox{0.78}{-} & \scalebox{0.78}{-} &\scalebox{0.78}{0.080} &\scalebox{0.78}{0.196} &  {\scalebox{0.78}{0.086}} &  {\scalebox{0.78}{0.206}} & \scalebox{0.78}{0.093} & \scalebox{0.78}{0.217} &  {\scalebox{0.78}{0.088}} &  {\scalebox{0.78}{0.205}} & \scalebox{0.78}{0.256} & \scalebox{0.78}{0.367} & \scalebox{0.78}{0.094} & \scalebox{0.78}{0.218} & \scalebox{0.78}{0.107} & \scalebox{0.78}{0.234} & \scalebox{0.78}{0.088} & \scalebox{0.78}{0.218} & \scalebox{0.78}{0.267} & \scalebox{0.78}{0.396} & \scalebox{0.78}{0.148} & \scalebox{0.78}{0.278} & \scalebox{0.78}{0.111} & \scalebox{0.78}{0.237} & \scalebox{0.78}{0.197} & \scalebox{0.78}{0.323} \\ 
    &  \scalebox{0.78}{192}  &\scalebox{0.78}{0.161} &\scalebox{0.78}{0.295} &  \scalebox{0.78}{0.187} &\scalebox{0.78}{0.343} &\scalebox{0.78}{-} & \scalebox{0.78}{-} &\scalebox{0.78}{0.166} &\scalebox{0.78}{0.288} & \scalebox{0.78}{0.177} &  {\scalebox{0.78}{0.299}} & \scalebox{0.78}{0.184} & \scalebox{0.78}{0.307} &  {\scalebox{0.78}{0.176}} &  {\scalebox{0.78}{0.299}} & \scalebox{0.78}{0.470} & \scalebox{0.78}{0.509} & \scalebox{0.78}{0.184} & \scalebox{0.78}{0.307} & \scalebox{0.78}{0.226} & \scalebox{0.78}{0.344} &  {\scalebox{0.78}{0.176}} & \scalebox{0.78}{0.315} & \scalebox{0.78}{0.351} & \scalebox{0.78}{0.459} & \scalebox{0.78}{0.271} & \scalebox{0.78}{0.315} & \scalebox{0.78}{0.219} & \scalebox{0.78}{0.335} & \scalebox{0.78}{0.300} & \scalebox{0.78}{0.369} \\ 
    &  \scalebox{0.78}{336}  &\scalebox{0.78}{0.309} &\scalebox{0.78}{0.333} &  \scalebox{0.78}{0.353} &\scalebox{0.78}{0.473} &\scalebox{0.78}{-} & \scalebox{0.78}{-} &\scalebox{0.78}{0.307} &\scalebox{0.78}{0.398} & \scalebox{0.78}{0.331} &  {\scalebox{0.78}{0.417}} & \scalebox{0.78}{0.351} & \scalebox{0.78}{0.432}&  {\scalebox{0.78}{0.301}} &  {\scalebox{0.78}{0.397}} & \scalebox{0.78}{1.268} & \scalebox{0.78}{0.883} & \scalebox{0.78}{0.349} & \scalebox{0.78}{0.431} & \scalebox{0.78}{0.367} & \scalebox{0.78}{0.448} &  {\scalebox{0.78}{0.313}} & \scalebox{0.78}{0.427} & \scalebox{0.78}{1.324} & \scalebox{0.78}{0.853} & \scalebox{0.78}{0.460} & \scalebox{0.78}{0.427} & \scalebox{0.78}{0.421} & \scalebox{0.78}{0.476} & \scalebox{0.78}{0.509} & \scalebox{0.78}{0.524} \\ 
    &  \scalebox{0.78}{720}  &\scalebox{0.78}{0.632} &\scalebox{0.78}{0.617} &  \scalebox{0.78}{0.934} &\scalebox{0.78}{0.761} &\scalebox{0.78}{-} & \scalebox{0.78}{-} &\scalebox{0.78}{0.656} &\scalebox{0.78}{0.582} &  {\scalebox{0.78}{0.847}} &  {\scalebox{0.78}{0.691}} & \scalebox{0.78}{0.886} & \scalebox{0.78}{0.714} & \scalebox{0.78}{0.901} & \scalebox{0.78}{0.714} & \scalebox{0.78}{1.767} & \scalebox{0.78}{1.068} & \scalebox{0.78}{0.852} & \scalebox{0.78}{0.698} & \scalebox{0.78}{0.964} & \scalebox{0.78}{0.746} &  {\scalebox{0.78}{0.839}} & \scalebox{0.78}{0.695} & \scalebox{0.78}{1.058} & \scalebox{0.78}{0.797} & \scalebox{0.78}{1.195} &  {\scalebox{0.78}{0.695}} & \scalebox{0.78}{1.092} & \scalebox{0.78}{0.769} & \scalebox{0.78}{1.447} & \scalebox{0.78}{0.941} \\ 
    \cmidrule(lr){2-32}
    &  \scalebox{0.78}{Avg}  &\scalebox{0.78}{\boldres{0.275}} &\scalebox{0.78}{\secondres{0.361}} &  \scalebox{0.78}{0.391} &\scalebox{0.78}{0.453} &\scalebox{0.78}{-} & \scalebox{0.78}{-} &\scalebox{0.78}{0.302} &\scalebox{0.78}{0.366} &  {\scalebox{0.78}{0.360}} &  {\scalebox{0.78}{0.403}} & \scalebox{0.78}{0.378} & \scalebox{0.78}{0.417} & \scalebox{0.78}{0.367} &  {\scalebox{0.78}{0.404}} & \scalebox{0.78}{0.940} & \scalebox{0.78}{0.707} & \scalebox{0.78}{0.370} & \scalebox{0.78}{0.413} & \scalebox{0.78}{0.416} & \scalebox{0.78}{0.443} &  {\scalebox{0.78}{0.354}} & \scalebox{0.78}{0.414} & \scalebox{0.78}{0.750} & \scalebox{0.78}{0.626} & \scalebox{0.78}{0.519} & \scalebox{0.78}{0.429} & \scalebox{0.78}{0.461} & \scalebox{0.78}{0.454} & \scalebox{0.78}{0.613} & \scalebox{0.78}{0.539} \\

    \midrule
    
    \multirow{5}{*}{\rotatebox{90}{\scalebox{0.95}{Traffic}}} 
    & \scalebox{0.78}{96}  &\scalebox{0.78}{0.372} &\scalebox{0.78}{0.241} &  \scalebox{0.78}{0.462} &\scalebox{0.78}{0.285} &\scalebox{0.78}{0.468} & \scalebox{0.78}{0.268} &\scalebox{0.78}{0.368} &\scalebox{0.78}{0.253} &  {\scalebox{0.78}{0.395}} &  {\scalebox{0.78}{0.268}} & \scalebox{0.78}{0.649} & \scalebox{0.78}{0.389} &  {\scalebox{0.78}{0.462}} & \scalebox{0.78}{0.295} & \scalebox{0.78}{0.522} &  {\scalebox{0.78}{0.290}} & \scalebox{0.78}{0.805} & \scalebox{0.78}{0.493} &{\scalebox{0.78}{0.593}} &{\scalebox{0.78}{0.321}} &\scalebox{0.78}{0.650} &\scalebox{0.78}{0.396} & \scalebox{0.78}{0.788} & \scalebox{0.78}{0.499} &{\scalebox{0.78}{0.587}} &\scalebox{0.78}{0.366} &\scalebox{0.78}{0.612} &{\scalebox{0.78}{0.338}} &\scalebox{0.78}{0.613} &\scalebox{0.78}{0.388} \\ 
    & \scalebox{0.78}{192}  &\scalebox{0.78}{0.384} &\scalebox{0.78}{0.259} &  \scalebox{0.78}{0.473} &\scalebox{0.78}{0.296} &\scalebox{0.78}{0.413} & \scalebox{0.78}{0.317} &\scalebox{0.78}{0.379} &\scalebox{0.78}{0.261} &  {\scalebox{0.78}{0.417}} &  {\scalebox{0.78}{0.276}} & \scalebox{0.78}{0.601} & \scalebox{0.78}{0.366} &  {\scalebox{0.78}{0.466}} & \scalebox{0.78}{0.296} & \scalebox{0.78}{0.530} &  {\scalebox{0.78}{0.293}} & \scalebox{0.78}{0.756} & \scalebox{0.78}{0.474} &\scalebox{0.78}{0.617} &{\scalebox{0.78}{0.336}} &{\scalebox{0.78}{0.598}} &\scalebox{0.78}{0.370} & \scalebox{0.78}{0.789} & \scalebox{0.78}{0.505} &\scalebox{0.78}{0.604} &\scalebox{0.78}{0.373} &\scalebox{0.78}{0.613} &{\scalebox{0.78}{0.340}} &\scalebox{0.78}{0.616} &\scalebox{0.78}{0.382}  \\ 
    & \scalebox{0.78}{336}  &\scalebox{0.78}{0.395} &\scalebox{0.78}{0.257} &  \scalebox{0.78}{0.498} &\scalebox{0.78}{0.296} &\scalebox{0.78}{0.529} & \scalebox{0.78}{0.284} &\scalebox{0.78}{0.397} &\scalebox{0.78}{0.270} &  {\scalebox{0.78}{0.433}} &  {\scalebox{0.78}{0.283}} & \scalebox{0.78}{0.609} & \scalebox{0.78}{0.369} &  {\scalebox{0.78}{0.482}} &  {\scalebox{0.78}{0.304}} & \scalebox{0.78}{0.558} & \scalebox{0.78}{0.305}  & \scalebox{0.78}{0.762} & \scalebox{0.78}{0.477} &\scalebox{0.78}{0.629} &{\scalebox{0.78}{0.336}}  &{\scalebox{0.78}{0.605}} &\scalebox{0.78}{0.373} & \scalebox{0.78}{0.797} & \scalebox{0.78}{0.508}&\scalebox{0.78}{0.621} &\scalebox{0.78}{0.383} &\scalebox{0.78}{0.618} &{\scalebox{0.78}{0.328}} &\scalebox{0.78}{0.622} &\scalebox{0.78}{0.337} \\ 
    & \scalebox{0.78}{720}&\scalebox{0.78}{0.458} &\scalebox{0.78}{0.302} &  \scalebox{0.78}{0.506} &\scalebox{0.78}{0.313} &\scalebox{0.78}{0.564} & \scalebox{0.78}{0.297} &\scalebox{0.78}{0.440} &\scalebox{0.78}{0.296} &  {\scalebox{0.78}{0.467}} &  {\scalebox{0.78}{0.302}} & \scalebox{0.78}{0.647} & \scalebox{0.78}{0.387} &  {\scalebox{0.78}{0.514}} &  {\scalebox{0.78}{0.322}} & \scalebox{0.78}{0.589} & \scalebox{0.78}{0.328}  & \scalebox{0.78}{0.719} & \scalebox{0.78}{0.449} &\scalebox{0.78}{0.640} &{\scalebox{0.78}{0.350}} &\scalebox{0.78}{0.645} &\scalebox{0.78}{0.394} & \scalebox{0.78}{0.841} & \scalebox{0.78}{0.523} &{\scalebox{0.78}{0.626}} &\scalebox{0.78}{0.382} &\scalebox{0.78}{0.653} &{\scalebox{0.78}{0.355}} &\scalebox{0.78}{0.660} &\scalebox{0.78}{0.408} \\ 
    \cmidrule(lr){2-32}
    & \scalebox{0.78}{Avg}  &\scalebox{0.78}{\secondres{0.401}} &\scalebox{0.78}{\boldres{0.269}} &  \scalebox{0.78}{0.484} &\scalebox{0.78}{0.297} &\scalebox{0.78}{0.493} & \scalebox{0.78}{0.291} &\scalebox{0.78}{\boldres{0.398}} &\scalebox{0.78}{\secondres{0.270}} &  {\scalebox{0.78}{0.428}} &  {\scalebox{0.78}{0.282}} & \scalebox{0.78}{0.626} & \scalebox{0.78}{0.378} &  {\scalebox{0.78}{0.481}} &  {\scalebox{0.78}{0.304}}& \scalebox{0.78}{0.550} &  {\scalebox{0.78}{0.304}} & \scalebox{0.78}{0.760} & \scalebox{0.78}{0.473} &{\scalebox{0.78}{0.620}} &{\scalebox{0.78}{0.336}} &\scalebox{0.78}{0.625} &\scalebox{0.78}{0.383} & \scalebox{0.78}{0.804} & \scalebox{0.78}{0.509} &{\scalebox{0.78}{0.610}} &\scalebox{0.78}{0.376} &\scalebox{0.78}{0.624} &{\scalebox{0.78}{0.340}} &\scalebox{0.78}{0.628} &\scalebox{0.78}{0.379} \\ 
    \midrule
    
    \multirow{5}{*}{\rotatebox{90}{\scalebox{0.95}{Weather}}} 
    &  \scalebox{0.78}{96}  &\scalebox{0.78}{0.150} &\scalebox{0.78}{0.201} &  \scalebox{0.78}{0.163} &\scalebox{0.78}{0.209} &\scalebox{0.78}{0.176} & \scalebox{0.78}{0.219} &\scalebox{0.78}{0.149} &\scalebox{0.78}{0.200} & \scalebox{0.78}{0.174} &  {\scalebox{0.78}{0.214}} & \scalebox{0.78}{0.192} & \scalebox{0.78}{0.232} & \scalebox{0.78}{0.177} &  {\scalebox{0.78}{0.218}} &  {\scalebox{0.78}{0.158}} & \scalebox{0.78}{0.230}  & \scalebox{0.78}{0.202} & \scalebox{0.78}{0.261} & {\scalebox{0.78}{0.172}} &{\scalebox{0.78}{0.220}} & \scalebox{0.78}{0.196} &\scalebox{0.78}{0.255} & \scalebox{0.78}{0.221} & \scalebox{0.78}{0.306} & \scalebox{0.78}{0.217} &\scalebox{0.78}{0.296} & {\scalebox{0.78}{0.173}} &{\scalebox{0.78}{0.223}} & \scalebox{0.78}{0.266} &\scalebox{0.78}{0.336} \\ 
    & \scalebox{0.78}{192} &\scalebox{0.78}{0.167} &\scalebox{0.78}{0.236} &  \scalebox{0.78}{0.222} &\scalebox{0.78}{0.260} &\scalebox{0.78}{0.222} & \scalebox{0.78}{0.260} &\scalebox{0.78}{0.196} &\scalebox{0.78}{0.245} & \scalebox{0.78}{0.221} &  {\scalebox{0.78}{0.254}} & \scalebox{0.78}{0.240} & \scalebox{0.78}{0.271} & \scalebox{0.78}{0.225} & \scalebox{0.78}{0.259} &  {\scalebox{0.78}{0.206}} & \scalebox{0.78}{0.277} & \scalebox{0.78}{0.242} & \scalebox{0.78}{0.298} & {\scalebox{0.78}{0.219}} & {\scalebox{0.78}{0.261}}  & \scalebox{0.78}{0.237} &\scalebox{0.78}{0.296} & \scalebox{0.78}{0.261} & \scalebox{0.78}{0.340} & \scalebox{0.78}{0.276} &\scalebox{0.78}{0.336} & \scalebox{0.78}{0.245} &\scalebox{0.78}{0.285} & \scalebox{0.78}{0.307} &\scalebox{0.78}{0.367} \\ 
    & \scalebox{0.78}{336}  &\scalebox{0.78}{0.224} &\scalebox{0.78}{0.257} &  \scalebox{0.78}{0.251} &\scalebox{0.78}{0.287} &\scalebox{0.78}{0.275} & \scalebox{0.78}{0.297} &\scalebox{0.78}{0.238} &\scalebox{0.78}{0.277} &  {\scalebox{0.78}{0.278}} &  {\scalebox{0.78}{0.296}} & \scalebox{0.78}{0.292} & \scalebox{0.78}{0.307} & \scalebox{0.78}{0.278} &  {\scalebox{0.78}{0.297}} &  {\scalebox{0.78}{0.272}} & \scalebox{0.78}{0.335} & \scalebox{0.78}{0.287} & \scalebox{0.78}{0.335} &{\scalebox{0.78}{0.280}} &{\scalebox{0.78}{0.306}} & \scalebox{0.78}{0.283} &\scalebox{0.78}{0.335} & \scalebox{0.78}{0.309} & \scalebox{0.78}{0.378} & \scalebox{0.78}{0.339} &\scalebox{0.78}{0.380} & \scalebox{0.78}{0.321} &\scalebox{0.78}{0.338} & \scalebox{0.78}{0.359} &\scalebox{0.78}{0.395}\\ 
    & \scalebox{0.78}{720}  &\scalebox{0.78}{0.298} &\scalebox{0.78}{0.302} &  \scalebox{0.78}{0.350} &\scalebox{0.78}{0.349} &\scalebox{0.78}{0.350} & \scalebox{0.78}{0.349} &\scalebox{0.78}{0.314} &\scalebox{0.78}{0.334} & \scalebox{0.78}{0.358} & {\scalebox{0.78}{0.347}} & \scalebox{0.78}{0.364} & \scalebox{0.78}{0.353} & \scalebox{0.78}{0.354} &  {\scalebox{0.78}{0.348}} & \scalebox{0.78}{0.398} & \scalebox{0.78}{0.418} &  {\scalebox{0.78}{0.351}} & \scalebox{0.78}{0.366} &\scalebox{0.78}{0.365} &{\scalebox{0.78}{0.359}} &  {\scalebox{0.78}{0.345}} &{\scalebox{0.78}{0.381}} & \scalebox{0.78}{0.377} & \scalebox{0.78}{0.427} & \scalebox{0.78}{0.403} &\scalebox{0.78}{0.428} & \scalebox{0.78}{0.414} &\scalebox{0.78}{0.410} & \scalebox{0.78}{0.419} &\scalebox{0.78}{0.428} \\ 
    \cmidrule(lr){2-32}
    & \scalebox{0.78}{Avg}  &\scalebox{0.78}{{\boldres{0.210}}} &\scalebox{0.78}{\boldres{0.257}} &  \scalebox{0.78}{0.240} &\scalebox{0.78}{0.271} &\scalebox{0.78}{0.255} & \scalebox{0.78}{0.280} &\scalebox{0.78}{\secondres{0.224}} &\scalebox{0.78}{\secondres{0.264}} &  {\scalebox{0.78}{0.258}} &  {\scalebox{0.78}{0.278}} & \scalebox{0.78}{0.272} & \scalebox{0.78}{0.291} &  {\scalebox{0.78}{0.259}} &  {\scalebox{0.78}{0.281}} & \scalebox{0.78}{0.259} & \scalebox{0.78}{0.315} & \scalebox{0.78}{0.271} & \scalebox{0.78}{0.320} &{\scalebox{0.78}{0.259}} &{\scalebox{0.78}{0.287}} &\scalebox{0.78}{0.265} &\scalebox{0.78}{0.317} & \scalebox{0.78}{0.292} & \scalebox{0.78}{0.363} &\scalebox{0.78}{0.309} &\scalebox{0.78}{0.360} &\scalebox{0.78}{0.288} &\scalebox{0.78}{0.314} &\scalebox{0.78}{0.338} &\scalebox{0.78}{0.382} \\ 
    \midrule
    
        \multirow{5}{*}{\rotatebox{90}{\scalebox{0.95}{ECL}}} 
    &  \scalebox{0.78}{96} & \scalebox{0.78}{0.132} &\scalebox{0.78}{0.235} &\scalebox{0.78}{0.153} &\scalebox{0.78}{0.247} &  \scalebox{0.78}{0.165} &\scalebox{0.78}{0.253} &\scalebox{0.78}{0.129} &\scalebox{0.78}{0.226} &  {\scalebox{0.78}{0.148}} &  {\scalebox{0.78}{0.240}} & \scalebox{0.78}{0.201} & \scalebox{0.78}{0.281} & \scalebox{0.78}{0.181} &  {\scalebox{0.78}{0.270}} & \scalebox{0.78}{0.219} & \scalebox{0.78}{0.314} & \scalebox{0.78}{0.237} & \scalebox{0.78}{0.329} & {\scalebox{0.78}{0.168}} &\scalebox{0.78}{0.272} &\scalebox{0.78}{0.197} &\scalebox{0.78}{0.282} & \scalebox{0.78}{0.247} & \scalebox{0.78}{0.345} &\scalebox{0.78}{0.193} &\scalebox{0.78}{0.308} &{\scalebox{0.78}{0.169}} &{\scalebox{0.78}{0.273}} &\scalebox{0.78}{0.201} &\scalebox{0.78}{0.317}  \\ 
    & \scalebox{0.78}{192} & \scalebox{0.78}{0.141} &\scalebox{0.78}{0.219} &\scalebox{0.78}{0.166} &\scalebox{0.78}{0.256} &  \scalebox{0.78}{0.173} &\scalebox{0.78}{0.262} &\scalebox{0.78}{0.143} &\scalebox{0.78}{0.239} &  {\scalebox{0.78}{0.162}} &  {\scalebox{0.78}{0.253}} & \scalebox{0.78}{0.201} & \scalebox{0.78}{0.283} & \scalebox{0.78}{0.188} &  {\scalebox{0.78}{0.274}} & \scalebox{0.78}{0.231} & \scalebox{0.78}{0.322} & \scalebox{0.78}{0.236} & \scalebox{0.78}{0.330} &{\scalebox{0.78}{0.184}} &\scalebox{0.78}{0.289} &\scalebox{0.78}{0.196} &{\scalebox{0.78}{0.285}} & \scalebox{0.78}{0.257} & \scalebox{0.78}{0.355} &\scalebox{0.78}{0.201} &\scalebox{0.78}{0.315} & {\scalebox{0.78}{0.182}} &\scalebox{0.78}{0.286} &\scalebox{0.78}{0.222} &\scalebox{0.78}{0.334} \\ 
    & \scalebox{0.78}{336} & \scalebox{0.78}{0.157} &\scalebox{0.78}{0.252} &\scalebox{0.78}{0.185} &\scalebox{0.78}{0.277} &  \scalebox{0.78}{0.188} &\scalebox{0.78}{0.277} &\scalebox{0.78}{0.161} &\scalebox{0.78}{0.259} &  {\scalebox{0.78}{0.178}} &  {\scalebox{0.78}{0.269}} & \scalebox{0.78}{0.215} & \scalebox{0.78}{0.298} & \scalebox{0.78}{0.204} &  {\scalebox{0.78}{0.293}} & \scalebox{0.78}{0.246} & \scalebox{0.78}{0.337} & \scalebox{0.78}{0.249} & \scalebox{0.78}{0.344} & {\scalebox{0.78}{0.198}} &{\scalebox{0.78}{0.300}} &\scalebox{0.78}{0.209} &{\scalebox{0.78}{0.301}} & \scalebox{0.78}{0.269} & \scalebox{0.78}{0.369} &\scalebox{0.78}{0.214} &\scalebox{0.78}{0.329} &{\scalebox{0.78}{0.200}} &\scalebox{0.78}{0.304} &\scalebox{0.78}{0.231} &\scalebox{0.78}{0.338}  \\ 
    & \scalebox{0.78}{720} &\scalebox{0.78}{0.159} &\scalebox{0.78}{0.258} &\scalebox{0.78}{0.225} &\scalebox{0.78}{0.310} &  \scalebox{0.78}{0.214} &\scalebox{0.78}{0.305} &\scalebox{0.78}{0.191} &\scalebox{0.78}{0.286} &  {\scalebox{0.78}{0.225}} &  {\scalebox{0.78}{0.317}} & \scalebox{0.78}{0.257} & \scalebox{0.78}{0.331} & \scalebox{0.78}{0.246} & \scalebox{0.78}{0.324} & \scalebox{0.78}{0.280} & \scalebox{0.78}{0.363} & \scalebox{0.78}{0.284} & \scalebox{0.78}{0.373} & {\scalebox{0.78}{0.220}} & {\scalebox{0.78}{0.320}} &\scalebox{0.78}{0.245} &\scalebox{0.78}{0.333} & \scalebox{0.78}{0.299} & \scalebox{0.78}{0.390} &\scalebox{0.78}{0.246} &\scalebox{0.78}{0.355} &{\scalebox{0.78}{0.222}} &{\scalebox{0.78}{0.321}} &\scalebox{0.78}{0.254} &\scalebox{0.78}{0.361} \\ 
    \cmidrule(lr){2-32}
    & \scalebox{0.78}{Avg} & \scalebox{0.78}{\boldres{0.147}} &\scalebox{0.78}{\boldres{0.244}} &\scalebox{0.78}{0.182} &\scalebox{0.78}{0.272} &  \scalebox{0.78}{0.185} &\scalebox{0.78}{0.274} &\scalebox{0.78}{\secondres{0.156}} &\scalebox{0.78}{\secondres{0.253}} &  {\scalebox{0.78}{0.178}} &  {\scalebox{0.78}{0.270}} & \scalebox{0.78}{0.219} & \scalebox{0.78}{0.298} & \scalebox{0.78}{0.205} &  {\scalebox{0.78}{0.290}} & \scalebox{0.78}{0.244} & \scalebox{0.78}{0.334} & \scalebox{0.78}{0.251} & \scalebox{0.78}{0.344} & {\scalebox{0.78}{0.192}} &\scalebox{0.78}{0.295} &\scalebox{0.78}{0.212} &\scalebox{0.78}{0.300} & \scalebox{0.78}{0.268} & \scalebox{0.78}{0.365} &\scalebox{0.78}{0.214} &\scalebox{0.78}{0.327} &{\scalebox{0.78}{0.193}} &{\scalebox{0.78}{0.296}} &\scalebox{0.78}{0.227} &\scalebox{0.78}{0.338} \\ 
    \bottomrule
  \end{tabular}
    \end{small}
}
\end{table*}

\subsection{Time Series Classification}
The results are reported in \tabref{tab:full_classification_results}.

\begin{table*}[tbp]
  \caption{Full results for the classification task (accuracy \%). Best performance is shown in \boldres{bold}, and the second-best is \secondres{underlined}.}
  \label{tab:full_classification_results}
  \centering
  \resizebox{\textwidth}{!}{%
  \begin{tabular}{lccccccccccccccccccc}
    \toprule
    \multirow{2}{*}{\scalebox{0.78}{Datasets / Models}}   & \scalebox{0.6}{LSTM} & \scalebox{0.6}{LSTNet} & \scalebox{0.6}{LSSL} & \scalebox{0.7}{Trans.} & \scalebox{0.7}{Re.} & \scalebox{0.7}{In.} & \scalebox{0.7}{Pyra.} & \scalebox{0.7}{Auto.} & \scalebox{0.78}{Station.} &  \scalebox{0.7}{FED.} & \scalebox{0.7}{{/ETS.}} & \scalebox{0.7}{/Flow.} & \scalebox{0.7}{/DLinear} & \scalebox{0.7}{/LightTS.} &  \scalebox{0.7}{/TimesNet} & \scalebox{0.7}{/PatchTST/} & \scalebox{0.7}{MTCN/}  & \scalebox{0.75}{\textbf{\model}} \\
	&   \scalebox{0.7}{\citeyearpar{Hochreiter1997LongSM}} & 
	\scalebox{0.7}{\citeyearpar{2018Modeling}} & 
	\scalebox{0.7}{} & 
\scalebox{0.7}{\citeyearpar{vaswani2017attention}} & 
	\scalebox{0.7}{\citeyearpar{kitaev2020reformer}} & \scalebox{0.7}{\citeyearpar{zhou2021informer}} & \scalebox{0.7}{\citeyearpar{liu2021pyraformer}} &
	\scalebox{0.7}{\citeyearpar{wu2021autoformer}} & 
	\scalebox{0.7}{\citeyearpar{Liu2022NonstationaryTR}} &
	\scalebox{0.7}{\citeyearpar{zhou2022fedformer}} & \scalebox{0.7}{\citeyearpar{woo2022etsformer}} & \scalebox{0.7}{\citeyearpar{wu2022flowformer}} & 
	\scalebox{0.7}{\citeyearpar{Zeng2022AreTE}} & \scalebox{0.7}{\citeyearpar{Zhang2022LessIM}} & \scalebox{0.7}{\citeyearpar{wu2023timesnet}} & \scalebox{0.7}{\citeyearpar{nie2023a}} & \scalebox{0.7}{\citeyearpar{donghao2024moderntcn}}  & \scalebox{0.7}{(\textcolor{c1}{ours})} \\
    \midrule
	\scalebox{0.7}{EthanolConcentration} &   \scalebox{0.78}{32.3} & \scalebox{0.78}{39.9} & \scalebox{0.78}{31.1} & \scalebox{0.78}{32.7} &\scalebox{0.78}{31.9} &\scalebox{0.78}{31.6}   &\scalebox{0.78}{30.8} &\scalebox{0.78}{31.6} &\scalebox{0.78}{32.7} &\scalebox{0.78}{31.2} & \scalebox{0.78}{28.1} & \scalebox{0.78}{33.8} & \scalebox{0.78}{32.6} &\scalebox{0.78}{29.7} & \scalebox{0.78}{35.7} & \scalebox{0.78}{32.8} & \scalebox{0.78}{\secondres{36.3}} & \scalebox{0.78}{\boldres{39.0}} \\
	\scalebox{0.7}{FaceDetection} &  \scalebox{0.78}{57.7} & \scalebox{0.78}{65.7} & \scalebox{0.78}{66.7} & \scalebox{0.78}{67.3} & \scalebox{0.78}{68.6} &\scalebox{0.78}{67.0} &\scalebox{0.78}{65.7} &\scalebox{0.78}{68.4} &\scalebox{0.78}{68.0} &\scalebox{0.78}{66.0} & \scalebox{0.78}{66.3} & \scalebox{0.78}{67.6} &\scalebox{0.78}{68.0} &\scalebox{0.78}{67.5} & \scalebox{0.78}{68.6}  & \scalebox{0.78}{68.3} & \scalebox{0.78}{\secondres{70.8}} & \scalebox{0.78}{\boldres{71.1}} \\
	\scalebox{0.7}{Handwriting} &  \scalebox{0.78}{15.2} & \scalebox{0.78}{25.8} & \scalebox{0.78}{24.6} & \scalebox{0.78}{32.0} & \scalebox{0.78}{27.4} &\scalebox{0.78}{32.8} &\scalebox{0.78}{29.4} &\scalebox{0.78}{36.7} &\scalebox{0.78}{31.6} &\scalebox{0.78}{28.0} &  \scalebox{0.78}{32.5} & \scalebox{0.78}{33.8} & \scalebox{0.78}{27.0} &\scalebox{0.78}{26.1} & \scalebox{0.78}{32.1} & \scalebox{0.78}{29.6} & \scalebox{0.78}{\secondres{30.6}} & \scalebox{0.78}{\boldres{33.0}} \\
	\scalebox{0.7}{Heartbeat} & \scalebox{0.78}{72.2} & \scalebox{0.78}{77.1} & \scalebox{0.78}{72.7} & \scalebox{0.78}{76.1} & \scalebox{0.78}{77.1} &\scalebox{0.78}{80.5} &\scalebox{0.78}{75.6} &\scalebox{0.78}{74.6} &\scalebox{0.78}{73.7} &\scalebox{0.78}{73.7} &  \scalebox{0.78}{71.2} & \scalebox{0.78}{77.6} & \scalebox{0.78}{75.1} &\scalebox{0.78}{75.1} & \scalebox{0.78}{78.0} & \scalebox{0.78}{74.9} & \scalebox{0.78}{\secondres{77.2}} & \scalebox{0.78}{\boldres{78.4}}     \\
	\scalebox{0.7}{JapaneseVowels}  & \scalebox{0.78}{79.7} & \scalebox{0.78}{98.1} & \scalebox{0.78}{98.4}  & \scalebox{0.78}{98.7} & \scalebox{0.78}{97.8} &\scalebox{0.78}{98.9} &\scalebox{0.78}{98.4} &\scalebox{0.78}{96.2} &\scalebox{0.78}{99.2} &\scalebox{0.78}{98.4} & \scalebox{0.78}{95.9} &  \scalebox{0.78}{98.9} & \scalebox{0.78}{96.2} &\scalebox{0.78}{96.2} & \scalebox{0.78}{98.4}  & \scalebox{0.78}{97.5} & \scalebox{0.78}{\secondres{98.8}} & \scalebox{0.78}{\boldres{98.9}} \\
	\scalebox{0.7}{PEMS-SF} & \scalebox{0.78}{39.9} & \scalebox{0.78}{86.7} & \scalebox{0.78}{86.1} & \scalebox{0.78}{82.1} & \scalebox{0.78}{82.7} &\scalebox{0.78}{81.5} &\scalebox{0.78}{83.2} &\scalebox{0.78}{82.7} &\scalebox{0.78}{87.3} &\scalebox{0.78}{80.9} & \scalebox{0.78}{86.0} &  \scalebox{0.78}{83.8} & \scalebox{0.78}{75.1} &\scalebox{0.78}{88.4} & \scalebox{0.78}{89.6} & \scalebox{0.78}{89.3} & \scalebox{0.78}{\secondres{89.1}} & \scalebox{0.78}{\boldres{90.0}} \\
	\scalebox{0.7}{SelfRegulationSCP1} & \scalebox{0.78}{68.9} & \scalebox{0.78}{84.0} & \scalebox{0.78}{90.8} & \scalebox{0.78}{92.2} & \scalebox{0.78}{90.4} &\scalebox{0.78}{90.1} &\scalebox{0.78}{88.1} &\scalebox{0.78}{84.0} &\scalebox{0.78}{89.4} &\scalebox{0.78}{88.7} & \scalebox{0.78}{89.6} & \scalebox{0.78}{92.5} & \scalebox{0.78}{87.3} &\scalebox{0.78}{89.8} & \scalebox{0.78}{91.8}  & \scalebox{0.78}{90.7} & \scalebox{0.78}{\secondres{93.4}} & \scalebox{0.78}{\boldres{94.2}} \\
    \scalebox{0.7}{SelfRegulationSCP2} & \scalebox{0.78}{46.6} & \scalebox{0.78}{52.8} & \scalebox{0.78}{52.2}  & \scalebox{0.78}{53.9} & \scalebox{0.78}{56.7} &\scalebox{0.78}{53.3} &\scalebox{0.78}{53.3} &\scalebox{0.78}{50.6} &\scalebox{0.78}{57.2} &\scalebox{0.78}{54.4} & \scalebox{0.78}{55.0} &  \scalebox{0.78}{56.1} & \scalebox{0.78}{50.5} &\scalebox{0.78}{51.1} & \scalebox{0.78}{57.2} & \scalebox{0.78}{57.8} & \scalebox{0.78}{\secondres{60.3}} & \scalebox{0.78}{\boldres{62.0}}  \\
    \scalebox{0.7}{SpokenArabicDigits} & \scalebox{0.78}{31.9} & \scalebox{0.78}{100.0} & \scalebox{0.78}{100.0}  & \scalebox{0.78}{98.4} & \scalebox{0.78}{97.0} &\scalebox{0.78}{100.0} &\scalebox{0.78}{\secondres{99.6}} &\scalebox{0.78}{100.0} &\scalebox{0.78}{100.0} &\scalebox{0.78}{100.0} & \scalebox{0.78}{100.0} &  \scalebox{0.78}{98.8} & \scalebox{0.78}{81.4} &\scalebox{0.78}{\boldres{100.0}} & \scalebox{0.78}{99.0} & \scalebox{0.78}{98.3} & \scalebox{0.78}{98.7} & \scalebox{0.78}{98.6} \\
    \scalebox{0.7}{UWaveGestureLibrary}  & \scalebox{0.78}{41.2} & \scalebox{0.78}{87.8} & \scalebox{0.78}{85.9}  & \scalebox{0.78}{85.6} & \scalebox{0.78}{85.6} &\scalebox{0.78}{85.6} &\scalebox{0.78}{83.4} &\scalebox{0.78}{85.9} &\scalebox{0.78}{87.5} &\scalebox{0.78}{85.3} & \scalebox{0.78}{85.0} &  \scalebox{0.78}{86.6} & \scalebox{0.78}{82.1} &\scalebox{0.78}{80.3} & \scalebox{0.78}{85.3} & \scalebox{0.78}{85.8} & \scalebox{0.78}{\secondres{86.7}} & \scalebox{0.78}{\boldres{88.3}} \\
    \midrule
    \scalebox{0.78}{Average Accuracy} &  \scalebox{0.78}{48.6} & \scalebox{0.78}{71.8} & \scalebox{0.78}{70.9}  & \scalebox{0.78}{71.9} & \scalebox{0.78}{71.5} &\scalebox{0.78}{72.1} &\scalebox{0.78}{70.8} &\scalebox{0.78}{71.1} &\scalebox{0.78}{72.7} &\scalebox{0.78}{70.7} & \scalebox{0.78}{71.0} &    {\scalebox{0.78}{73.0}} & \scalebox{0.78}{67.5} &\scalebox{0.78}{70.4} &  {\scalebox{0.78}{73.6}} & \scalebox{0.78}{72.5} & \scalebox{0.78}{\secondres{74.2}} & \scalebox{0.78}{\boldres{75.68}} \\
    \bottomrule
  \end{tabular}%
  }
\end{table*}

\subsection{Time Series Anomaly Detection}
The results are reported in \tabref{tab:full_anomaly_results}.

\begin{table*}[tbp]
  \caption{Full results for the anomaly detection task. The P, R and F1 represent the precision, recall and F1-score in percentage respectively. A higher value of P, R and F1 indicates a better performance. Best performance is shown in \boldres{bold}, and the second-best is \secondres{underlined}. We take the average of 5 separate runs for each dataset.}
  \label{tab:full_anomaly_results}
  \vskip 0.05in
  \centering
  \begin{small}
  \renewcommand{\multirowsetup}{\centering}
  \setlength{\tabcolsep}{1.4pt}
  \begin{tabular}{lc|ccc|ccc|ccc|ccc|ccc|c}
    \toprule
    \multicolumn{2}{c}{\scalebox{0.8}{Datasets}} & 
    \multicolumn{3}{c}{\scalebox{0.8}{\rotatebox{0}{SMD}}} &
    \multicolumn{3}{c}{\scalebox{0.8}{\rotatebox{0}{MSL}}} &
    \multicolumn{3}{c}{\scalebox{0.8}{\rotatebox{0}{SMAP}}} &
    \multicolumn{3}{c}{\scalebox{0.8}{\rotatebox{0}{SWaT}}} & 
    \multicolumn{3}{c}{\scalebox{0.8}{\rotatebox{0}{PSM}}} & \scalebox{0.8}{Avg F1} \\
    \cmidrule(lr){3-5} \cmidrule(lr){6-8} \cmidrule(lr){9-11} \cmidrule(lr){12-14} \cmidrule(lr){15-17} \cmidrule(lr){18-18}
    \multicolumn{2}{c}{\scalebox{0.8}{Metrics}} & \scalebox{0.8}{P} & \scalebox{0.8}{R} & \scalebox{0.8}{F1} 
    & \scalebox{0.8}{P} & \scalebox{0.8}{R} & \scalebox{0.8}{F1} 
    & \scalebox{0.8}{P} & \scalebox{0.8}{R} & \scalebox{0.8}{F1} 
    & \scalebox{0.8}{P} & \scalebox{0.8}{R} & \scalebox{0.8}{F1} 
    & \scalebox{0.8}{P} & \scalebox{0.8}{R} & \scalebox{0.8}{F1} 
    & \scalebox{0.8}{(\%)}\\
    \midrule
        \scalebox{0.85}{LSTM} &
        \scalebox{0.85}{\citeyearpar{Hochreiter1997LongSM}}      
        & \scalebox{0.85}{78.52} & \scalebox{0.85}{65.47} & \scalebox{0.85}{71.41} 
        & \scalebox{0.85}{78.04} & \scalebox{0.85}{86.22} & \scalebox{0.85}{81.93}
        & \scalebox{0.85}{91.06} & \scalebox{0.85}{57.49} & \scalebox{0.85}{70.48} 
        & \scalebox{0.85}{78.06} & \scalebox{0.85}{91.72} & \scalebox{0.85}{84.34} 
        & \scalebox{0.85}{69.24} & \scalebox{0.85}{\secondres{99.53}} & \scalebox{0.85}{81.67}
        & \scalebox{0.85}{77.97} \\ 
        \scalebox{0.85}{Transformer} &
        \scalebox{0.85}{\citeyearpar{vaswani2017attention}}    
        & \scalebox{0.85}{83.58} & \scalebox{0.85}{76.13} & \scalebox{0.85}{79.56} 
        & \scalebox{0.85}{71.57} & \scalebox{0.85}{87.37} & \scalebox{0.85}{78.68}
        & \scalebox{0.85}{89.37} & \scalebox{0.85}{57.12} & \scalebox{0.85}{69.70} 
        & \scalebox{0.85}{68.84} & \scalebox{0.85}{96.53} & \scalebox{0.85}{80.37}
        & \scalebox{0.85}{62.75} & \scalebox{0.85}{96.56} & \scalebox{0.85}{76.07}
        & \scalebox{0.85}{76.88} \\ 
        \scalebox{0.85}{LogTrans} & \scalebox{0.85}{\citeyearpar{2019Enhancing}}
        & \scalebox{0.85}{83.46} & \scalebox{0.85}{70.13} & \scalebox{0.85}{76.21} 
        & \scalebox{0.85}{73.05} & \scalebox{0.85}{87.37} & \scalebox{0.85}{79.57}
        & \scalebox{0.85}{89.15} & \scalebox{0.85}{57.59} & \scalebox{0.85}{69.97} 
        & \scalebox{0.85}{68.67} & \scalebox{0.85}{97.32} & \scalebox{0.85}{80.52}
        & \scalebox{0.85}{63.06} & \scalebox{0.85}{98.00} & \scalebox{0.85}{76.74}
        & \scalebox{0.85}{76.60} \\ 
        \scalebox{0.85}{TCN} & 
        \scalebox{0.85}{\citeyearpar{Franceschi2019UnsupervisedSR}} 
        & \scalebox{0.85}{84.06} & \scalebox{0.85}{79.07} & \scalebox{0.85}{81.49} 
        & \scalebox{0.85}{75.11} & \scalebox{0.85}{82.44} & \scalebox{0.85}{78.60}
        & \scalebox{0.85}{86.90} & \scalebox{0.85}{\boldres{59.23}} & \scalebox{0.85}{70.45} 
        & \scalebox{0.85}{76.59} & \scalebox{0.85}{95.71} & \scalebox{0.85}{85.09}
        & \scalebox{0.85}{54.59} & \scalebox{0.85}{\boldres{99.77}} & \scalebox{0.85}{70.57}
        & \scalebox{0.85}{77.24} \\
        \scalebox{0.85}{Reformer} & \scalebox{0.85}{\citeyearpar{kitaev2020reformer}}
        & \scalebox{0.85}{82.58} & \scalebox{0.85}{69.24} & \scalebox{0.85}{75.32} 
        & \scalebox{0.85}{\boldres{85.51}} & \scalebox{0.85}{83.31} & \scalebox{0.85}{84.40}
        & \scalebox{0.85}{90.91} & \scalebox{0.85}{57.44} & \scalebox{0.85}{70.40} 
        & \scalebox{0.85}{72.50} & \scalebox{0.85}{96.53} & \scalebox{0.85}{82.80}
        & \scalebox{0.85}{59.93} & \scalebox{0.85}{95.38} & \scalebox{0.85}{73.61}
        & \scalebox{0.85}{77.31} \\ 
        \scalebox{0.85}{Informer} & \scalebox{0.85}{\citeyearpar{zhou2021informer}}
        & \scalebox{0.85}{86.60} & \scalebox{0.85}{77.23} & \scalebox{0.85}{81.65} 
        & \scalebox{0.85}{81.77} & \scalebox{0.85}{86.48} & \scalebox{0.85}{84.06}
        & \scalebox{0.85}{90.11} & \scalebox{0.85}{57.13} & \scalebox{0.85}{69.92} 
        & \scalebox{0.85}{70.29} & \scalebox{0.85}{96.75} & \scalebox{0.85}{81.43}
        & \scalebox{0.85}{64.27} & \scalebox{0.85}{96.33} & \scalebox{0.85}{77.10}
        & \scalebox{0.85}{78.83} \\ 
        \scalebox{0.85}{Anomaly$^\ast$} & \scalebox{0.85}{\citeyearpar{xu2021anomaly}} 
        & \scalebox{0.85}{\boldres{88.91}} & \scalebox{0.85}{82.23} & \scalebox{0.85}{85.49}
        & \scalebox{0.85}{79.61} & \scalebox{0.85}{87.37} & \scalebox{0.85}{83.31}
        & \scalebox{0.85}{91.85} & \scalebox{0.85}{58.11} & \scalebox{0.85}{71.18}
        & \scalebox{0.85}{72.51} & \scalebox{0.85}{97.32} & \scalebox{0.85}{83.10}
        & \scalebox{0.85}{68.35} & \scalebox{0.85}{94.72} & \scalebox{0.85}{79.40}
        & \scalebox{0.85}{80.50} \\
        \scalebox{0.85}{Pyraformer} & \scalebox{0.85}{\citeyearpar{liu2021pyraformer}}
        & \scalebox{0.85}{85.61} & \scalebox{0.85}{80.61} & \scalebox{0.85}{83.04} 
        & \scalebox{0.85}{83.81} & \scalebox{0.85}{85.93} & \scalebox{0.85}{84.86}
        & \scalebox{0.85}{92.54} & \scalebox{0.85}{57.71} & \scalebox{0.85}{71.09}
        & \scalebox{0.85}{87.92} & \scalebox{0.85}{96.00} & \scalebox{0.85}{91.78}
        & \scalebox{0.85}{71.67} & \scalebox{0.85}{96.02} & \scalebox{0.85}{82.08}
        & \scalebox{0.85}{82.57} \\
        \scalebox{0.85}{Autoformer} & \scalebox{0.85}{\citeyearpar{wu2021autoformer}}
        & \scalebox{0.85}{88.06} & \scalebox{0.85}{82.35} & \scalebox{0.85}{85.11}
        & \scalebox{0.85}{77.27} & \scalebox{0.85}{80.92} & \scalebox{0.85}{79.05}
        & \scalebox{0.85}{90.40} & \scalebox{0.85}{58.62} & \scalebox{0.85}{71.12}
        & \scalebox{0.85}{89.85} & \scalebox{0.85}{95.81} & \scalebox{0.85}{92.74}
        & \scalebox{0.85}{99.08} & \scalebox{0.85}{88.15} & \scalebox{0.85}{93.29}
        & \scalebox{0.85}{84.26} \\
        \scalebox{0.85}{Stationary} & \scalebox{0.85}{\citeyearpar{Liu2022NonstationaryTR}}
        & \scalebox{0.85}{88.33} & \scalebox{0.85}{81.21} & \scalebox{0.85}{84.62}
        & \scalebox{0.85}{68.55} & \scalebox{0.85}{\secondres{89.14}} & \scalebox{0.85}{77.50}
        & \scalebox{0.85}{89.37} & \scalebox{0.85}{\secondres{59.02}} & \scalebox{0.85}{71.09}
        & \scalebox{0.85}{68.03} & \scalebox{0.85}{96.75} & \scalebox{0.85}{79.88}
        & \scalebox{0.85}{97.82} & \scalebox{0.85}{96.76} & \scalebox{0.85}{\secondres{97.29}}
        & \scalebox{0.85}{82.08} \\
        \scalebox{0.85}{DLinear} & \scalebox{0.85}{\citeyearpar{Zeng2022AreTE}}
        & \scalebox{0.85}{83.62} & \scalebox{0.85}{71.52} & \scalebox{0.85}{77.10}
        & \scalebox{0.85}{84.34} & \scalebox{0.85}{85.42} & \scalebox{0.85}{84.88}
        & \scalebox{0.85}{92.32} & \scalebox{0.85}{55.41} & \scalebox{0.85}{69.26}
        & \scalebox{0.85}{80.91} & \scalebox{0.85}{95.30} & \scalebox{0.85}{87.52}
        & \scalebox{0.85}{98.28} & \scalebox{0.85}{89.26} & \scalebox{0.85}{93.55}
        & \scalebox{0.85}{82.46} \\
        \scalebox{0.85}{{ETSformer}} & \scalebox{0.85}{\citeyearpar{woo2022etsformer}}
        & \scalebox{0.85}{87.44} & \scalebox{0.85}{79.23} & \scalebox{0.85}{83.13}
        & \scalebox{0.85}{\secondres{85.13}} & \scalebox{0.85}{84.93} & \scalebox{0.85}{85.03}
        & \scalebox{0.85}{92.25} & \scalebox{0.85}{55.75} & \scalebox{0.85}{69.50}
        & \scalebox{0.85}{90.02} & \scalebox{0.85}{80.36} & \scalebox{0.85}{84.91}
        & \scalebox{0.85}{\boldres{99.31}} & \scalebox{0.85}{85.28} & \scalebox{0.85}{91.76}
        & \scalebox{0.85}{82.87} \\
        \scalebox{0.85}{LightTS} & \scalebox{0.85}{\citeyearpar{Zhang2022LessIM}}
        & \scalebox{0.85}{87.10} & \scalebox{0.85}{78.42} & \scalebox{0.85}{82.53}
        & \scalebox{0.85}{82.40} & \scalebox{0.85}{75.78} & \scalebox{0.85}{78.95}
        & \scalebox{0.85}{92.58} & \scalebox{0.85}{55.27} & \scalebox{0.85}{69.21}
        & \scalebox{0.85}{\secondres{91.98}} & \scalebox{0.85}{94.72} & \scalebox{0.85}{93.33}
        & \scalebox{0.85}{98.37} & \scalebox{0.85}{95.97} & \scalebox{0.85}{97.15}
        & \scalebox{0.85}{84.23} \\
        \scalebox{0.85}{FEDformer} & \scalebox{0.85}{\citeyearpar{zhou2022fedformer}}
        & \scalebox{0.85}{87.95} & \scalebox{0.85}{82.39} & \scalebox{0.85}{85.08}
        & \scalebox{0.85}{77.14} & \scalebox{0.85}{80.07} & \scalebox{0.85}{78.57}
        & \scalebox{0.85}{90.47} & \scalebox{0.85}{58.10} & \scalebox{0.85}{70.76}
        & \scalebox{0.85}{90.17} & \scalebox{0.85}{96.42} & \scalebox{0.85}{93.19}
        & \scalebox{0.85}{97.31} & \scalebox{0.85}{97.16} & \scalebox{0.85}{97.23}
        & \scalebox{0.85}{84.97} \\
        \scalebox{0.85}{TimesNet (I)} & \scalebox{0.85}{\citeyearpar{wu2023timesnet}}
        & \scalebox{0.85}{87.76} & \scalebox{0.85}{82.63} & \scalebox{0.85}{85.12}
        & \scalebox{0.85}{82.97} & \scalebox{0.85}{85.42} & \scalebox{0.85}{84.18}
        & \scalebox{0.85}{91.50} & \scalebox{0.85}{57.80} & \scalebox{0.85}{70.85}
        & \scalebox{0.85}{88.31} & \scalebox{0.85}{96.24} & \scalebox{0.85}{92.10}
        & \scalebox{0.85}{98.22} & \scalebox{0.85}{92.21} & \scalebox{0.85}{95.21}
        & \scalebox{0.85}{85.49} \\
        \scalebox{0.85}{TimesNet (R)} & \scalebox{0.85}{\citeyearpar{wu2023timesnet}}
        & \scalebox{0.85}{\secondres{88.66}} & \scalebox{0.85}{83.14} & \scalebox{0.85}{\secondres{85.81}}
        & \scalebox{0.85}{83.92} & \scalebox{0.85}{86.42} & \scalebox{0.85}{\secondres{85.15}}
        & \scalebox{0.85}{92.52} & \scalebox{0.85}{58.29} & \scalebox{0.85}{\boldres{71.52}}
        & \scalebox{0.85}{86.76} & \scalebox{0.85}{\boldres{97.32}} & \scalebox{0.85}{91.74}
        & \scalebox{0.85}{98.19} & \scalebox{0.85}{96.76} & \scalebox{0.85}{\boldres{97.47}}
        & \scalebox{0.85}{86.34} \\
        \scalebox{0.85}{CrossFormer} & \scalebox{0.85}{\citeyearpar{zhang2022crossformer}}
        & \scalebox{0.85}{83.60} & \scalebox{0.85}{76.61} & \scalebox{0.85}{79.70}
        & \scalebox{0.85}{84.68} & \scalebox{0.85}{83.71} & \scalebox{0.85}{84.19}
        & \scalebox{0.85}{92.04} & \scalebox{0.85}{55.37} & \scalebox{0.85}{69.14}
        & \scalebox{0.85}{88.49} & \scalebox{0.85}{93.48} & \scalebox{0.85}{90.92}
        & \scalebox{0.85}{97.16} & \scalebox{0.85}{89.73} & \scalebox{0.85}{93.30}
        & \scalebox{0.85}{83.45} \\
        \scalebox{0.85}{PatchTST} & \scalebox{0.85}{\citeyearpar{nie2023a}}
        & \scalebox{0.85}{87.42} & \scalebox{0.85}{81.65} & \scalebox{0.85}{84.44}
        & \scalebox{0.85}{84.07} & \scalebox{0.85}{86.23} & \scalebox{0.85}{85.14}
        & \scalebox{0.85}{92.43} & \scalebox{0.85}{57.51} & \scalebox{0.85}{70.91}
        & \scalebox{0.85}{80.70} & \scalebox{0.85}{94.93} & \scalebox{0.85}{87.24}
        & \scalebox{0.85}{98.87} & \scalebox{0.85}{93.99} & \scalebox{0.85}{96.37}
        & \scalebox{0.85}{84.82} \\
        \scalebox{0.85}{ModernTCN} & \scalebox{0.85}{\citeyearpar{donghao2024moderntcn}}
        & \scalebox{0.85}{87.86} & \scalebox{0.85}{\secondres{83.85}} & \scalebox{0.85}{\secondres{85.81}}
        & \scalebox{0.85}{83.94} & \scalebox{0.85}{85.93} & \scalebox{0.85}{84.92}
        & \scalebox{0.85}{\secondres{93.17}} & \scalebox{0.85}{57.69} & \scalebox{0.85}{\secondres{71.26}}
        & \scalebox{0.85}{91.83} & \scalebox{0.85}{95.98} & \scalebox{0.85}{\secondres{93.86}}
        & \scalebox{0.85}{98.09} & \scalebox{0.85}{96.38} & \scalebox{0.85}{97.23}
        & \scalebox{0.85}{\secondres{86.62}} \\
        \scalebox{0.85}{\model{}} & \scalebox{0.85}{(\textcolor{c1}{ours})}
        & \scalebox{0.85}{88.34} & \scalebox{0.85}{\boldres{86.13}} & \scalebox{0.85}{\boldres{86.61}}
        & \scalebox{0.85}{83.73} & \scalebox{0.85}{\boldres{89.43}} & \scalebox{0.85}{\boldres{86.21}}
        & \scalebox{0.85}{\boldres{93.33}} & \scalebox{0.85}{57.85} & \scalebox{0.85}{70.90}
        & \scalebox{0.85}{\boldres{92.27}} & \scalebox{0.85}{\secondres{96.77}} & \scalebox{0.85}{\boldres{94.22}}
        & \scalebox{0.85}{\secondres{99.20}} & \scalebox{0.85}{94.56} & \scalebox{0.85}{96.79}
        & \scalebox{0.85}{\boldres{87.66}} \\
    \bottomrule
  \end{tabular}
  \end{small}
  \vspace{-10pt}
\end{table*}

\section{Illustrative Example}
To concretely illustrate the operation of \model{}, we construct a small synthetic example: a multivariate time series with three variates and six time steps,
\begin{align*}
\mathbf{X} & =
\begin{pmatrix}
x_{1,1} & x_{1,2} & x_{1,3} & x_{1,4} & x_{1,5} & x_{1,6} \\
x_{2,1} & x_{2,2} & x_{2,3} & x_{2,4} & x_{2,5} & x_{2,6} \\
x_{3,1} & x_{3,2} & x_{3,3} & x_{3,4} & x_{3,5} & x_{3,6}
\end{pmatrix}
\\=&
\begin{pmatrix}
1 & 2 & 3 & 4 & 5 & 6 \\
2 & 3 & 4 & 5 & 6 & 7 \\
3 & 4 & 5 & 6 & 7 & 8
\end{pmatrix},
\end{align*}

chosen for its arithmetic regularity, which enables the exact, closed-form computation of all intermediate values. Following the conventions introduced in \secref{sec:hydra}, we adopt scalar key and value projections: \(k_{t,v} = x_{t,v}\) and \(v_{t,v} = 2x_{t,v}\). The memory states of both associative heads are initialized in the logarithmic domain as \(\tilde{\M}^{(1)}_{0,v} = \tilde{\M}^{(2)}_{0,v} = 0\), resulting in multiplicative memory weights \(\M^{(1)}_{0,v} = \M^{(2)}_{0,v} = 1\).
To isolate architectural behavior, all gating parameters are fixed to \(\alpha = \beta = \theta = \mu = 0.5\), while all learning rates are uniformly set to \(\eta = \gamma = \lambda = \omega = 10^{-2}\). Computation proceeds in raster-scan order: time steps are traversed from left to right, and within each time step, variates are processed top to bottom.

At each location \((t,v)\), the first memory head is updated according to
\[
\tilde{\M}^{(1)}_{t,v} = \alpha\,\tilde{\M}^{(1)}_{t-1,v} - \eta\,g^{(1)}_{t,v} + \beta\,\tilde{\M}^{(2)}_{t-1,v} - \gamma\,g^{(2)}_{t,v},
\]
where \(g^{(i)}_{t,v} = \left(\M^{(i)}_{t-1,v}k_{t,v} - v_{t,v}\right)k_{t,v}\) denotes the prediction error of head \(i\) based on its previous state. The second head is updated similarly, but along the variate axis:
\[
\tilde{\M}^{(2)}_{t,v} = \theta\,\tilde{\M}^{(1)}_{t,v-1} - \lambda\,\tilde{g}^{(1)}_{t,v} + \mu\,\tilde{\M}^{(2)}_{t,v-1} - \omega\,\tilde{g}^{(2)}_{t,v},
\]
where \(\tilde{g}^{(i)}_{t,v}\) denotes the prediction error of head \(i\) computed at the previous variate. After each step, the logarithmic memory states are exponentiated to yield the actual associative weights propagated forward.
We walk through every \((t,v)\) in raster order, explicitly computing
\[
(g^{(1)}_{t,v},\,g^{(2)}_{t,v})
\;\to\;
(\tilde{\M}^{(1)}_{t,v},\,\tilde{\M}^{(2)}_{t,v})
\;\to\;
(\M^{(1)}_{t,v},\,\M^{(2)}_{t,v}).
\]
\noindent
Below we detail the sequential computation of update values for first few $(t,v)$ entries.
\allowdisplaybreaks
\begin{footnotesize}
\begin{flalign*}
& \textbf{(1,1):}\\
& g^{(1)}_{1,1} = ((1.000\cdot1 - 2)\cdot1) = -1.000\\
& g^{(2)}_{1,1} = ((1.000\cdot1 - 2)\cdot1) = -1.000\\
& \tilde{\M}^{(1)}_{1,1} = 0.5\cdot0.000 \\
& -0.01(-1.000) \\
& +0.5\cdot0.000 \\
& -0.01(-1.000) = 0.02000\\
& \tilde{\M}^{(2)}_{1,1} = 0.5\cdot0.000 \\
& -0.01(-1.000) \\
& +0.5\cdot0.000 \\
& -0.01(-1.000) = 0.02000\\
& \M^{(1)}_{1,1} \approx e^{0.02000} = 1.020\\
& \M^{(2)}_{1,1} \approx e^{0.02000} = 1.020\\[1ex]
& \textbf{(1,2):}\\
& g^{(1)}_{1,2} = ((1.020\cdot2 - 4)\cdot2) = -3.919\\
& g^{(2)}_{1,2} = ((1.020\cdot2 - 4)\cdot2) = -3.919\\
& \tilde{\M}^{(1)}_{1,2} = 0.5\cdot0.02000 \\
& -0.01(-3.919) \\
& +0.5\cdot0.02000 \\
& -0.01(-3.919) = 0.09838\\
& \tilde{\M}^{(2)}_{1,2} = 0.5\cdot0.00000 \\
& -0.01(-3.919) \\
& +0.5\cdot0.00000 \\
& -0.01(-3.919) = 0.07838\\
& \M^{(1)}_{1,2} \approx e^{0.09838} = 1.103\\
& \M^{(2)}_{1,2} \approx e^{0.07838} = 1.082\\[1ex]
& \textbf{(1,3):}\\
& g^{(1)}_{1,3} = ((1.103\cdot3 - 6)\cdot3) = -8.070\\
& g^{(2)}_{1,3} = ((1.082\cdot3 - 6)\cdot3) = -8.266\\
& \tilde{\M}^{(1)}_{1,3} = 0.5\cdot0.09838 \\
& -0.01(-8.070) \\
& +0.5\cdot0.07838 \\
& -0.01(-8.266) = 0.25174\\
& \tilde{\M}^{(2)}_{1,3} = 0.5\cdot0.00000 \\
& -0.01(-8.070) \\
& +0.5\cdot0.00000 \\
& -0.01(-8.266) = 0.16336\\
& \M^{(1)}_{1,3} \approx e^{0.25174} = 1.286\\
& \M^{(2)}_{1,3} \approx e^{0.16336} = 1.177\\[1ex]
& \textbf{(1,4):}\\
& g^{(1)}_{1,4} = ((1.286\cdot4 - 8)\cdot4) = -11.420\\
& g^{(2)}_{1,4} = ((1.177\cdot4 - 8)\cdot4) = -13.161\\
& \tilde{\M}^{(1)}_{1,4} = 0.5\cdot0.25174 \\
& -0.01(-11.420) \\
& +0.5\cdot0.16336 \\
& -0.01(-13.161) = 0.45335\\
& \tilde{\M}^{(2)}_{1,4} = 0.5\cdot0.00000 \\
& -0.01(-11.420) \\
& +0.5\cdot0.00000 \\
& -0.01(-13.161) = 0.24580\\
& \M^{(1)}_{1,4} \approx e^{0.45335} = 1.573\\
& \M^{(2)}_{1,4} \approx e^{0.24580} = 1.279\\[1ex]
& \textbf{(1,5):}\\
& g^{(1)}_{1,5} = ((1.573\cdot5 - 10)\cdot5) = -10.660\\
& g^{(2)}_{1,5} = ((1.279\cdot5 - 10)\cdot5) = -18.034\\
& \tilde{\M}^{(1)}_{1,5} = 0.5\cdot0.45335 \\
& -0.01(-10.660) \\
& +0.5\cdot0.24580 \\
& -0.01(-18.034) = 0.63652\\
& \tilde{\M}^{(2)}_{1,5} = 0.5\cdot0.00000 \\
& -0.01(-10.660) \\
& +0.5\cdot0.00000 \\
& -0.01(-18.034) = 0.28694\\
& \M^{(1)}_{1,5} \approx e^{0.63652} = 1.891\\
& \M^{(2)}_{1,5} \approx e^{0.28694} = 1.332\\[1ex]
& \textbf{(1,6):}\\
& g^{(1)}_{1,6} = ((1.891\cdot6 - 12)\cdot6) = -3.964\\
& g^{(2)}_{1,6} = ((1.332\cdot6 - 12)\cdot6) = -24.035\\
& \tilde{\M}^{(1)}_{1,6} = 0.5\cdot0.63652 \\
& -0.01(-3.964) \\
& +0.5\cdot0.28694 \\
& -0.01(-24.035) = 0.74172\\
& \tilde{\M}^{(2)}_{1,6} = 0.5\cdot0.00000 \\
& -0.01(-3.964) \\
& +0.5\cdot0.00000 \\
& -0.01(-24.035) = 0.27999\\
& \M^{(1)}_{1,6} \approx e^{0.74172} = 2.100\\
& \M^{(2)}_{1,6} \approx e^{0.27999} = 1.323\\[1ex]
& \textbf{(2,1):}\\
& g^{(1)}_{2,1} = ((1\cdot2 - 4)\cdot2) = -4.000\\
& g^{(2)}_{2,1} = ((1\cdot2 - 4)\cdot2) = -4.000\\
& \tilde{\M}^{(1)}_{2,1} = 0.5\cdot0.00000 \\
& -0.01(-4.000) \\
& +0.5\cdot0.00000 \\
& -0.01(-4.000) = 0.08000\\
& \tilde{\M}^{(2)}_{2,1} = 0.5\cdot0.02000 \\
& -0.01(-4.000) \\
& +0.5\cdot0.02000 \\
& -0.01(-4.000) = 0.10000\\
& \M^{(1)}_{2,1} \approx e^{0.08000} = 1.083\\
& \M^{(2)}_{2,1} \approx e^{0.10000} = 1.105\\[1ex]
& \textbf{(2,2):}\\
& g^{(1)}_{2,2} = ((1.083\cdot3 - 6)\cdot3) = -8.250\\
& g^{(2)}_{2,2} = ((1.105\cdot3 - 6)\cdot3) = -8.053\\
& \tilde{\M}^{(1)}_{2,2} = 0.5\cdot0.08000 \\
& -0.01(-8.250) \\
& +0.5\cdot0.10000 \\
& -0.01(-8.053) = 0.25304\\
& \tilde{\M}^{(2)}_{2,2} = 0.5\cdot0.09838 \\
& -0.01(-8.250) \\
& +0.5\cdot0.07838 \\
& -0.01(-8.053) = 0.25142\\
& \M^{(1)}_{2,2} \approx e^{0.25304} = 1.288\\
& \M^{(2)}_{2,2} \approx e^{0.25142} = 1.286\\
& \vdots
\end{flalign*}
\end{footnotesize}

Applying this update scheme across the full \(3 \times 6\) input matrix results in:
\[
\M^{(1)}=
\begin{pmatrix}
1.020 & 1.103 & 1.286 & 1.574 & 1.890 & 2.100\\
1.083 & 1.288 & 1.617 & 1.949 & 2.058 & 2.077\\
1.197 & 1.590 & 1.957 & 2.026 & 2.020 & 2.023
\end{pmatrix},
\]
\[
\M^{(2)}=
\begin{pmatrix}
1.020 & 1.082 & 1.177 & 1.279 & 1.332 & 1.323\\
1.105 & 1.286 & 1.546 & 1.749 & 1.769 & 1.814\\
1.310 & 1.634 & 1.919 & 1.930 & 1.949 & 1.979
\end{pmatrix}.
\]
with values rounded to three decimal places. Despite the simplicity of this example, it clearly demonstrates the bidirectional interaction between the two memory heads. Errors encountered by the first head at time \(t{-}1\) are partially mitigated by concurrent corrections from the second head before influencing the next temporal update. Likewise, the second head propagates information laterally across variates, influencing subsequent updates to the first head.

This dynamic coupling realizes a two-dimensional associative memory that is strictly more expressive than a set of independent recurrent filters applied along either axis alone. Furthermore, the model maintains efficient training behavior due to the chunked processing strategy introduced in \secref{sec:hydra}. The temporal ridge evident in \(\M^{(1)}\) highlights regions of aligned temporal and cross-variate evidence, while attenuation in \(\M^{(2)}\) illustrates areas of disagreement between the two memory streams.

\section{Compute Resources} \label{app:compute}
In our experiments, we use 4 NVIDIA A6000 GPUs with 48GB RAM.

\end{document}